\documentclass{article}

\usepackage{arxiv}

\usepackage{amsmath}
\usepackage{amssymb}
\usepackage{mathtools}
\usepackage{amsthm}
\usepackage{multirow}

\usepackage[utf8]{inputenc} 
\usepackage[T1]{fontenc}    
\usepackage{hyperref}       
\usepackage{url}            
\usepackage{booktabs}       
\usepackage{amsfonts}       
\usepackage{nicefrac}       
\usepackage{microtype}      
\usepackage{cleveref}       
\usepackage{lipsum}         
\usepackage{graphicx}
\usepackage{natbib}
\usepackage{doi}

\usepackage[textsize=tiny]{todonotes}
\usepackage[inline]{enumitem}
\usepackage{subcaption}
\usepackage{wrapfig}
\usepackage[table]{xcolor}

\def\model{\text{DFTNet}}
\def\modelt{\text{DFTNet(T)}}
\def\modelg{\text{DFTNet(G)}}
\def\modelm{\text{DFTNet(M)}}

\title{DFTNet: Deep Fourier Tokenization and Routing Codebook for Universal and Generalizable Time Series Forecasting}

\usepackage{authblk}

\author[1]{Yicheng Lu\thanks{Equal contribution.}}

\author[2]{Shihao Tu\protect\footnotemark[1]}

\author[2]{Wenyue Ding}
\author[1]{Qingkai Ren}
\author[1]{Yujun Chen}
\author[1]{Yin Zhang}

\author[1]{Yang Yang\thanks{Corresponding author: \texttt{yangya@zju.edu.cn}}}

\affil[1]{Zhejiang University, Hangzhou, China}
\affil[2]{SUPCON Technology Co., Ltd., Hangzhou, China}

\renewcommand{\undertitle}{}
\renewcommand{\headeright}{}

\hypersetup{
pdftitle={DFTNet: Deep Fourier Tokenization and Routing Codebook for Universal and Generalizable Time Series Forecasting},
pdfsubject={cs.LG},
pdfauthor={Yicheng Lu, Shihao Tu, Wenyue Ding, Qingkai Ren, Yujun Chen, Yin Zhang, Yang Yang},
pdfkeywords={Time series forecasting， Tokenization， Vector quantization， Cross-domain generalization},
}

\date{}

\begin{document}
\maketitle

\begin{abstract}
	Universal and generalizable time series forecasting has become an increasingly important goal, as real-world applications often involve diverse domains with varying temporal characteristics and data distributions. 
    To address this challenge, methods such as vector quantization (VQ) learn compact, discrete representations within a constrained latent space. However, existing VQ-based methods still suffer from limited codebook utilization and a lack of effective temporal decomposition, which hinder their generalization in complex multi-domain scenarios.
    To address these problems, this paper proposes \model{}, a novel two-stage time series forecasting model. This model achieves universal forecasting and effective generalization across multiple domains through two core innovations: first, to extract independent and informative deep bases, a Deep Fourier Tokenization (DFT) module is introduced with orthogonality and consistency constraints; second, to improve codebook utilization and expressiveness, this paper introduces the Routing Codebook, which replaces traditional distance metrics with a learnable routing strategy. Experimental results demonstrate that \model{} outperforms baseline models in prediction accuracy on multiple public datasets, especially showing competitive generalization capabilities in cross-domain prediction tasks.
\end{abstract}


\section{Introduction}
The rapid advancement of data acquisition technologies has led to vast accumulations of time series data across industries. This accumulation has fueled progress in forecasting tasks critical to applications such as electricity consumption~\cite{tu2024powerpm}, traffic flow~\cite{chen2026stunet}, and weather prediction. 
Existing forecasting models are often tailored to specific domains, limiting their applicability in diverse real-world scenarios. This has brought increasing attention to universal and generalizable time series forecasting methods that aim to perform robustly across a wide range of domains.

The varying temporal characteristics and data distributions of diverse domains require models to learn robust and transferable representations~\cite{tu2024dmnet}. To achieve this, recent work has explored vector quantization (VQ)-based methods~\cite{Oord2017NeuralDR}, which often build upon variational autoencoders (VAEs)~\cite{kingma2013auto} and employ a discrete codebook to map continuous input sequences into compact and discrete latent representations. 
Each codeword within this codebook signifies a representative pattern inherent in the time series. By quantizing continuous time series representations into codewords, these methods achieve information compression and redundancy reduction~\cite{Desai2021TimeVAEAV}, thus yielding efficient inputs.

Recent works highlight the potential of vector quantization tokens in time seires~\cite{Talukder2024TOTEMTT, Rasul2024VQTRVQ, shibo2025hdt}. TOTEM~\cite{Talukder2024TOTEMTT} demonstrated the robust generalization capability of vector quantization tokens for cross-domain time series forecasting. 
Meanwhile, HDT~\cite{shibo2025hdt} reframed time series prediction as a discrete token generation task. 

Despite the fact that these VQ-based models have achieved notable advancements in specific tasks, they still face several critical problems in practical applications. These problems primarily stem from the intrinsic complexity of time series and the inherent limitations in the design of current vector quantization methods:

\textbf{Lack of effective time series decomposition mechanisms. }
Time series exhibit complex long-range dependencies, multi-scale periodicities and non-stationary trends — unlike structured data (e.g., images and speech), where traditional vector quantization methods perform well~\cite{Razavi2019GeneratingDH, Chung2020VectorQuantizedAP, Ling2020DeCoAR2D, Gu2022RethinkingTO}. This coupling hinders the extraction of independent latent factors (deep bases), leading to redundancy and poor generalization. 

\textbf{Low codebook utilization due to distance-based codebook selection strategies. }
Current vector quantization methods typically rely on distance-based codebook selection strategies, often driven by Euclidean distance~\cite{Oord2017NeuralDR, Lee2023VectorQT, wen2024abstracted, Talukder2024TOTEMTT} or cosine similarity~\cite{Yu2021VectorquantizedIM, Chen2024SDformerSD}. These non-learnable nearest-neighbor selection mechanism struggle to adaptively map input representations to their most representative codewords when confronted with the diverse and unevenly distributed patterns in multi-domain data. Consequently, a small subset of codewords is frequently activated, whereas a substantial number remain underutilized, severely limiting the codebook expressiveness.

In light of the aforementioned challenges, we identify two critical challenges:
\begin{itemize}
\item 

\textbf{How to decouple time series into independent deep bases? }
The coupling of temporal patterns in time series makes it difficult to extract independent deep bases that are crucial for robust representation learning. Consequently, effectively decoupling these entangled temporal components and extract generalizable structured bases has become a pressing challenge.
\item

\textbf{How to design a learnable codebook selection mechanism for higher utilization? }
The distance-based selection strategies employed by current VQ-VAE models struggle to effectively capture the diverse patterns present in multi-domain time series, leading to suboptimal codebook utilization and limited expressiveness. Consequently, developing a learnable selection strategy that can adaptively choose codewords based on data characteristics is paramount to improving both codebook utilization and expressiveness.
\end{itemize}

To address the aforementioned challenges, this paper presents \model{}, a novel framework designed to achieve universal forecasting with robust generalization capabilities. At its core, we propose \underline{D}eep \underline{F}ourier \underline{T}okenization (DFT), a time series decomposition approach inspired by the Discrete Fourier Transform (DFT).
In classical Fourier analysis, a signal is decomposed into a set of orthogonal sine and cosine waves, each independently representing a specific frequency component. While mathematically elegant, these individual components often lack sufficient representational density for complex forecasting tasks. To overcome this limitation, we extend the principle of orthogonal decomposition by decoupling time series into multiple orthogonal deep bases. By imposing orthogonality and consistency losses, the DFT module effectively extracts independent and informative representations from continuous signals, thereby surmounting the constraints of conventional decoupling methods.
Furthermore, to tackle the prevalent issue of low codebook utilization, we introduce the Routing Codebook. Unlike traditional strategies driven by fixed Euclidean or cosine similarities, our approach employs a learnable routing mechanism that adaptively selects the most representative codewords. This strategy significantly enhances both codebook utilization and the expressiveness of the learned representations.
We summarize our main contributions as follows.

\begin{itemize}
\item We propose \model{}, which employs orthogonality and consistency constraints to decouple time series into structured discrete representations, providing a robust foundation for downstream forecasting.
\item We introduce the Routing Codebook, a learnable routing strategy that achieves adaptive codeword matching to significantly enhance codebook utilization and expressiveness.
\item Extensive experiments on eight datasets demonstrate that \model{} consistently outperforms SOTA baselines in most benchmark tests, exhibiting competitive robustness and generalization in zero-shot tasks.
\end{itemize}

\begin{figure*}[h]
    \centering
    \includegraphics[width=1\linewidth]{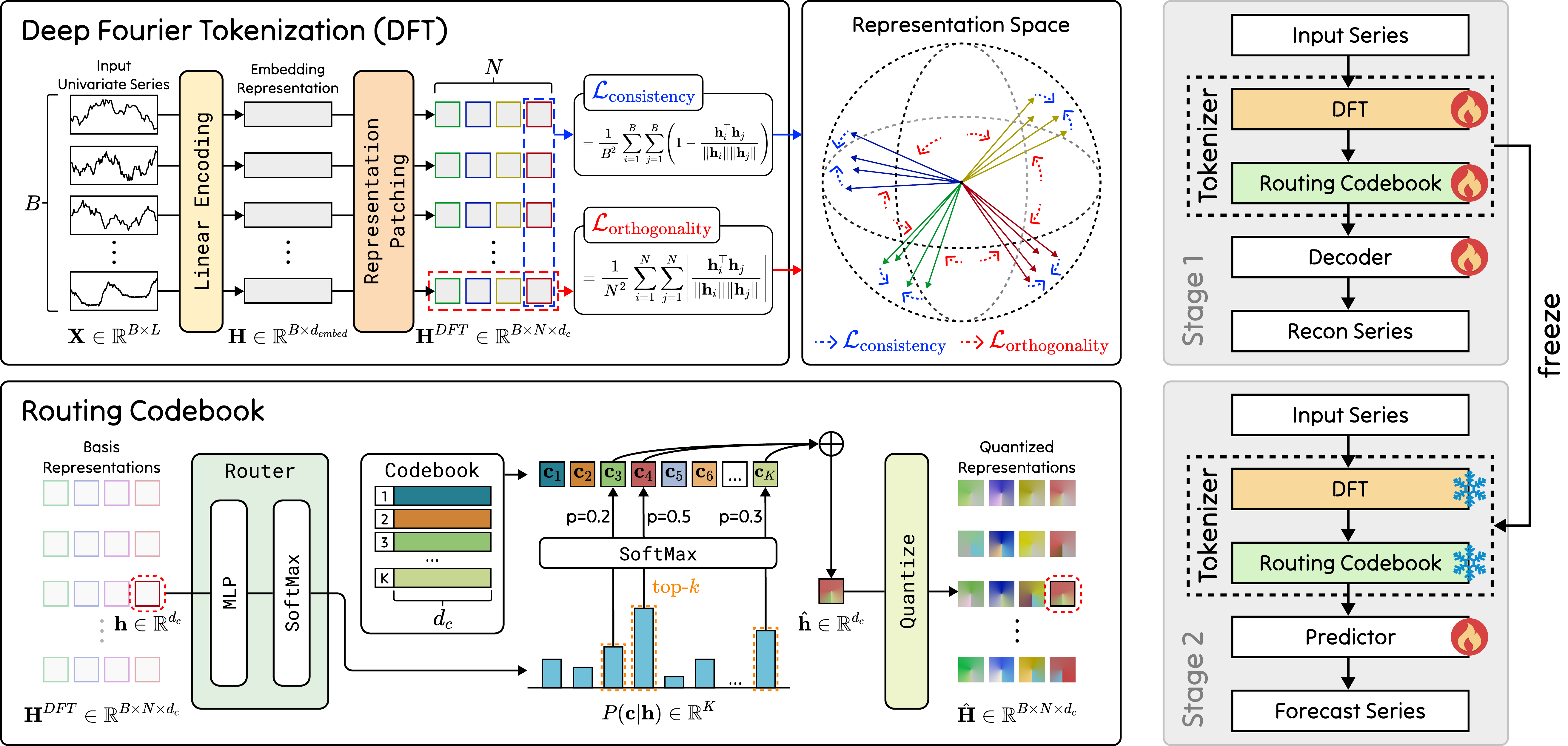}
    \caption{\model{} framework.}
    \vspace{-5pt}
    \label{fig:framework}
\end{figure*}

\section{Related works}

\subsection{Tokenization in Time Series}
Time series tokenization is the process of dividing a continuous numerical sequence into independent units or segments and mapping them to an embedding space. 
Existing time series tokenization methods can be broadly categorized into two main types: continuous representation-based tokenization (including point-based tokenization and patch-based tokenization) and discrete representation-based tokenization (including bin-based tokenization and vector quantization (VQ)-based tokenization).

\textbf{Continuous representation-based tokenization. }
Continuous representation-based tokenization maps time series to a continuous embedding space. Point-based tokenization, common in Transformer models~\cite{LI2019EnhancingTL, haoyietal-informer-2021, Wu2021AutoformerDT}, treats each time step as an independent token. While effective for long-term dependencies, it struggles with local feature extraction and high computational complexity. Patch-based tokenization~\cite{Yuqietal-2023-PatchTST, Zhang2023CrossformerTU} divides sequences into patches to reduce token count and improve efficiency, with each patch capturing richer local semantic information.

\textbf{Discrete representation-based tokenization. }
Discrete representation-based tokenization maps time series to a finite discrete space, enhancing generalization and abstraction. Bin-based tokenization discretizes numerical values into finite categories~\cite{lin2007experiencing, Rabanser2020TheEO}, while VQ-based tokenization maps embedding vectors to discrete categories in a learnable codebook~\cite{Oord2017NeuralDR, Desai2021TimeVAEAV, Lee2023VectorQT, Chen2024SDformerSD, wen2024abstracted}. Recent studies~\cite{Rasul2022VQARVQ, Talukder2024TOTEMTT, Rasul2024VQTRVQ, shibo2025hdt} highlight the generalization and pattern-capturing capabilities of VQ-based tokenization, with methods like TOTEM~\cite{Talukder2024TOTEMTT} using self-supervised learning and HDT~\cite{shibo2025hdt} applying hierarchical token generation for multivariate forecasting.

\subsection{Time Series Forecasting}
Time series forecasting is a core task in machine learning, aiming to infer future trends and dynamic structures from historical observation sequences. Early approaches such as ARIMA~\cite{kumar2015short} and ETS~\cite{gardner1985exponential} offer simple yet interpretable models, but struggle to capture complex temporal patterns and nonlinear dependencies.
To address these limitations, deep learning-based models have been developed, including recurrent architectures~\cite{rnn} like LSTM~\cite{hochreiter1997long} and GRU~\cite{Cho2014OnTP}, as well as convolution-based approaches~\cite{LeCun1998GradientbasedLA} such as TCN~\cite{BaiTCN2018}. These models provide improved capacity for modeling long-range and hierarchical temporal dependencies.

In recent years, the Transformer architecture~\cite{Vaswani2017AttentionIA} has been widely adopted in time series forecasting due to its strong modeling capabilities and scalability~\cite{haoyietal-informer-2021, Wu2021AutoformerDT, Zhou2022FEDformerFE, Yuqietal-2023-PatchTST, Zhang2023CrossformerTU, liu2023itransformer, tu2025astnet}. Models such as Informer~\cite{haoyietal-informer-2021} and Autoformer~\cite{Wu2021AutoformerDT} rely on point-based tokenization, whereas PatchTST~\cite{Yuqietal-2023-PatchTST} adopts patch-based tokenization, dividing long sequences into local patches, which improves modeling efficiency and the ability to capture local dynamics.

However, the high computational cost and complexity of these Transformer-based models have motivated the exploration of simpler and more efficient alternatives. In this context, a series of structurally simple linear models have emerged~\cite{Zeng2022AreTE, Li2023RevisitingLT, Das2023LongtermFW}. For example, DLinear~\cite{Zeng2022AreTE} explicitly models trend and seasonal components and operates directly on raw sequences or their decompositions, simplifying the modeling process. PatchMLP~\cite{Tang2025UnlockingTP} replaces attention with a patch-wise MLP structure, achieving competitive long-term forecasting performance with a much simpler design.

More recently, researchers have also begun exploring the construction of pre-trained foundation models, aiming to address multiple tasks such as forecasting, interpolation, and anomaly detection through a unified modeling framework. Representative works such as GPT4TS~\cite{Zhou2023OneFA}, Moirai~\cite{woo2024moirai}, Chronos~\cite{ansari2024chronos}, and Time-MOE~\cite{xiaoming2025time} leverage large-scale pre-training to improve cross-task and cross-domain generalization capabilities. 
Beyond forecasting, time-series foundation models have also been evaluated as simulators for process control~\cite{tu2026can}.

\section{Problem Formulation}
In univariate time series forecasting, given a historical sequence $\mathbf{X} = \{ \mathbf{x}_1,...,{\mathbf{x}}_L \} \in \mathbb{R}^{L}$ with $L$ time steps, we predict the future $T$ time steps $\mathbf{Y}=\{ \mathbf{x}_{L+1},...,{\mathbf{x}}_{L+T}\} \in \mathbb{R}^{T}$. Such an input sequence $\mathbf{X}$ and its corresponding target sequence $\mathbf{Y}$ constitute a complete sample pair. 
In this paper, we regard each dataset as an independent domain, focusing on two task scenarios:
\begin{enumerate*}[label=(\arabic*)]
    \item \textbf{In-domain Forecast}: The training and test samples are drawn from the same domain; and 
    \item \textbf{Cross-domain Forecast}: The training and test samples are drawn from different domains.
\end{enumerate*}

\section{Methodology}

\textbf{Overview.}
To tackle the challenges of complex temporal dependencies and low codebook utilization in vector quantization for time series forecasting, our method is designed as a two-stage framework (Figure~\ref{fig:framework}). The first stage focuses on learning structured discrete representations, which are then used for prediction in the second stage. 

\textbf{Stage 1: Discrete Representation Learning.}
\label{stage_1}
In the first stage, an encoder $\mathcal{E}(\cdot)$ maps the historical sequence $\mathbf{X}$ to an embedding representation $\mathbf{H}$. Subsequently, the Deep Fourier Tokenization (DFT) decomposes $\mathbf{H}$ into multiple basis categories and assembles them into a structured basis set $\mathbf{H}^{DFT}$. To achieve deeply decoupling bases, we impose orthogonality constraints on the bases decomposed from the same sequence, while simultaneously encouraging directional consistency among bases of different samples within the same category. 
The distinct basis categories represent latent universal features of time series, akin to classical components like frequency, amplitude, or seasonality. Unlike predefined features, these representations are autonomously learned through orthogonality and consistency constraints, resulting in a structured basis set with improved generalization, enabling the model to capture core temporal patterns across domains.
Then $\mathbf{H}^{DFT}$ are discretely quantized into discrete tokens $\hat{\mathbf{H}}$ using a routing mechanism-driven codebook $\mathcal{C}$. Finally, these discrete tokens are fed into a decoder $\mathcal{D}(\cdot)$ to obtain the reconstructed sequence $\hat{\mathbf{X}}$. The general formulation for the first stage can be expressed as follows.
\begin{align}
\begin{gathered}
\mathbf{H} = \mathcal{E} ( \mathbf{X} ) \\
\mathbf{H}^{DFT} =\text{DFT} ( \mathbf{H} ) \\
\hat{\mathbf{H}} = \text{RoutingCodebook} ( \mathbf{H}^{DFT} ) \\
\hat{\mathbf{X}} = \mathcal{D} ( \hat{\mathbf{H}} )
\end{gathered}
\end{align}

\textbf{Stage 2: Forecasting.}
\label{stage_2}
In the second stage, we freeze the parameters of the encoder and the codebook, treating them as a tokenizer to convert historical sequences into discrete tokens. These tokens serve as compact, expressive inputs to a downstream prediction model $\mathcal{F}(\cdot)$, which is trained to forecast the future sequence $\hat{\mathbf{Y}}$. This modular design facilitates flexible integration with diverse forecasting architectures while leveraging the rich, decoupled temporal representations learned in Stage ~\hyperref[stage_1]{1}. The general formulation for the second stage can be expressed as follows.
\begin{equation}
\begin{gathered}
\mathbf{H} = \mathcal{E}_{\text{frozen}} ( \mathbf{X} ) \\
\mathbf{H}^{DFT} =\text{DFT} ( \mathbf{H} ) \\
\hat{\mathbf{H}} = \text{RoutingCodebook}_{\text{frozen}} ( \mathbf{H}^{DFT} ) \\
\hat{\mathbf{Y}} =  \mathcal{F}( \hat{\mathbf{H}} )
\end{gathered}
\end{equation}

\subsection{Deep Fourier Tokenization (DFT)}
\label{DFT}
\textbf{Basis Decomposition.} 
To obtain informative and independent representations, we first decouple the time series into a finite set of bases.
Given a batch of $B$ historical sequence $\mathbf{X} \in \mathbb{R}^{B\times L}$, we feed it into an encoder $\mathcal{E}(\cdot)$ to obtain the embedding representation $\mathbf{H} = \mathcal{E} \left( \mathbf{X} \right) \in \mathbb{R}^{B\times d_{embed}}$, where $d_{embed}$ is the embedding dimension. Subsequently, we partition $\mathbf{H}$ into components of equal length along the dimension $d_{embed}$, resulting in: 
\begin{equation}
\begin{aligned}
\mathbf{H}^{DFT} &= \{ \mathbf{h}_{i,j} \mid i \in \{1, \dots, B\}, \\
&\quad j \in \{1, \dots, N\} \} \in \mathbb{R}^{B \times N \times d_{c}} 
\end{aligned}
\end{equation}
where $N \times d_{c} = d_{embed}$, with $N$ representing the number of basis categories, and $d_{c}$ is the dimension of each basis. 
We employed a strategy of encoding followed by splitting to generate tokens, and its effectiveness is validated in Appendix~\ref{app:additional_ablation_study_order_of_splitting_and_encoding}.
For convenience, we use $\mathbf{H}^{DFT}_{:n}$ to denote the bases originating from the same time series $i$ and $\mathbf{H}^{DFT}_{b:}$ to denote the bases at the same relative position $j$ across all series in the batch.

\textbf{Orthogonality loss and consistency loss.}
To promote independence and coherence among the decoupled bases, we introduce two auxiliary constraints: orthogonality loss and consistency loss. These constraints guide the encoder $\mathcal{E}(\cdot)$ to produce more structured and disentangled representations.

To encourage bases from the same series, $\mathbf{H}^{DFT}_{:n} = \{\mathbf{h}_{1}, \dots, \mathbf{h}_{N}\} \in \mathbb{R}^{N\times d_{c}}$, to be mutually independent, we define the orthogonality loss as the average of the absolute values of cosine similarity between all pairs of bases:
\begin{equation}
\label{loss:ortho}
\mathcal{L}_{\text{ortho}}=\frac{\lambda}{N^2} \sum_{i=1}^{N} \sum_{j=1}^{N} 
\left| 
\frac{\mathbf{h}_{i}^{\top} \mathbf{h}_{j}}{\|\mathbf{h}_{i}\| \|\mathbf{h}_{j}\|}
\right|
\end{equation}
where $\| \cdot \|$ denotes the L2 norm and $\lambda$ is the orthogonality loss weight. Minimizing $\mathcal{L}_{\text{ortho}}$ encourages these bases to be orthogonal, thus capturing diverse latent factors in the time series.

Meanwhile, to enhance consistency across bases within the same basis category, $\mathbf{H}^{DFT}_{b:} = \{\mathbf{h}_{1}, \dots, \mathbf{h}_{B}\} \in \mathbb{R}^{B\times d_{c}}$, we define the consistency loss as the average difference between $1$ and cosine similarity across all pairs of bases:
\begin{equation}
\label{loss:consis}
\mathcal{L}_{\text{consis}}=\frac{\alpha}{B^2} \sum_{i=1}^{B} \sum_{j=1}^{B} 
\left(
1 - \frac{\mathbf{h}_{i}^{\top} \mathbf{h}_{j}}{\|\mathbf{h}_{i}\| \|\mathbf{h}_{j}\|}
\right)
\end{equation}
where $\alpha$ is the consistency loss weight. Here, since our objective is to maximize the cosine similarity of bases of the same category, we use its difference from $1$ as the optimization target, i.e. minimizing $\mathcal{L}_{\text{consis}}$, to enhance consistency.
Together, $\mathcal{L}_{\text{ortho}}$ and $\mathcal{L}_{\text{consis}}$ enable the model to decouple time series into more structured, diverse, and semantically meaningful bases. These two objectives are added to the total optimization objective $\mathcal{L}$ in Section~\ref{routing_codebook}.

\subsection{Routing Codebook}
\label{routing_codebook}
\textbf{Routing mechanism-driven selection strategy.}
As described in Section~\ref{DFT}, the $\text{DFT}$ receives the embedding $\mathbf{H}$ and produces the bases $\mathbf{H}^{DFT}$ as input to the codebook $\mathcal{C}$. Then, for each basis $\mathbf{h} \in \mathbb{R}^{d_{c}}$ within $\mathbf{H}^{DFT}$, we select the $\text{top-}k$ most relevant codewords from a codebook $\mathcal{C} = \{\mathbf{c}_{1}, \dots, \mathbf{c}_{K}\} \in \mathbb{R}^{K \times d_{c}}$, which contains $K$ codewords of dimension $d_{c}$. These selected codewords are then combined via a weighted sum to form the quantized representation $\hat{\mathbf{H}}$.

To enable learnable codeword matching, we introduce a routing mechanism to guide the selection.
Specifically, for each basis $\mathbf{h}$, we employ a two-layer multilayer perceptron ($\text{MLP}$) to compute routing scores, which are then passed through a $\text{Softmax}(\cdot)$ function to yield a probability distribution over codewords:
\begin{equation}
\mathbf{P}(\mathbf{c} | \mathbf{h}) = \text{Softmax}(\mathbf{W}_2 ~ \sigma(\mathbf{W}_1 \mathbf{h} + b_1) + b_2) \in \mathbb{R}^{K}
\end{equation}
where $\sigma(\cdot)$ is an activation function, while $\mathbf{W}_i$ and $\mathbf{b}_i$ represent the weight matrix and bias of the $i\text{-th}$ layer of $\text{MLP}$, respectively. 
From $\mathcal{C}$, we select the top-$k$ codewords $\mathcal{C}_{\text{top-}k}$ with the highest probabilities $\mathbf{P}(\mathbf{c} | \mathbf{h})_{\text{top-}k}$ in $\mathbf{P}(\mathbf{c} | \mathbf{h})$. These probabilities are re-normalized using softmax to ensure they sum to $1$, and the quantized representation $\hat{\mathbf{h}}$ is computed as a weighted sum:
\begin{equation}
\hat{\mathbf{h}} = \sum_{i=1}^{\text{top-}k} \text{Softmax}(p_i) \mathbf{c}_i
\end{equation}
where $p_i$ and $\mathbf{c}_i$ are the $i\text{-th}$ elements of $\mathbf{P}(\mathbf{c} | \mathbf{h})_{\text{top-}k}$ and $\mathcal{C}_{\text{top-}k}$, respectively. Consequently, the overall quantized representation of $\mathbf{H}^{DFT}$ can be denoted as:
\begin{equation}
\hat{\mathbf{H}} = \{ \hat{\mathbf{h}}_{i,j} \mid i \in \{1, \dots, B\},~ j \in \{1, \dots, N\} \} \in \mathbb{R}^{B \times N \times d_{c}}
\end{equation}
Similarly, we use $\hat{\mathbf{H}}_{:n} = \{\hat{\mathbf{h}}_{1}, \dots, \hat{\mathbf{h}}_{N}\} \in \mathbb{R}^{N\times d_{c}}$
to denote the quantized representations of the same series. Finally, a decoder $\mathcal{D}(\cdot)$, which mirrors the architecture of the encoder, maps the flattened representation $\hat{\mathbf{H}}_{:n} \in \mathbb{R}^{(N \times d_{c})}$ to the reconstructed time series $\hat{\mathbf{X}} = \mathcal{D} ( \hat{\mathbf{H}}_{:n} ) \in \mathbb{R}^L$.

\textbf{Optimization objective.}
To train this time series tokenizer, we employ the following loss functions to optimize the parameters of the encoder $\mathcal{E}(\cdot)$, the decoder $\mathcal{D}(\cdot)$, and the $\text{MLP}$ for routing:
\begin{equation}
\begin{gathered}
\mathcal{L}_{\text{recon}} = \frac{1}{B} \sum_{i=1}^{B} \| \mathbf{X}_i - \hat{\mathbf{X}}_i \|_2^2 \\
\mathcal{L}_{\text{commit}} = \frac{\beta}{B \cdot N} \sum_{i=1}^{B \cdot N} \|\mathbf{h}_i - \operatorname{sg}[\hat{\mathbf{h}}_i]\|_2^2 \\
\mathcal{L} = \mathcal{L}_{\text{recon}} + \mathcal{L}_{\text{commit}} + \mathcal{L}_{\text{ortho}} + \mathcal{L}_{\text{consis}}
\end{gathered}
\end{equation}
The total loss $\mathcal{L}$ consists of four components. The first is reconstruction loss ($\mathcal{L}_{\text{recon}}$), which penalizes the difference between the original and reconstructed sequences. The second is commitment loss ($\mathcal{L}_{\text{commit}}$),which prevents the encoder output from being far from the embedding vectors (codewords), where $\operatorname{sg}[\cdot]$ denotes the stop-gradient operator and $\beta$ is the weighting coefficient for the commitment loss. The third and fourth loss terms correspond to the orthogonality loss ($\mathcal{L}_{\text{ortho}}$) and the consistency loss ($\mathcal{L}_{\text{consis}}$) in the Deep Fourier Tokenization, which were discussed in detail in Equation~\ref{loss:ortho} and Equation~\ref{loss:consis}. The codebook $\mathcal{C}$ is updated using Exponential Moving Average (EMA).

Once the first training stage is complete, we freeze the parameters of the encoder $\mathcal{E}(\cdot)$, the codebook $\mathcal{C}$, and the $\text{MLP}$ for routing, collectively function as the time series tokenizer. With this tokenizer, a univariate time series $\mathbf{X} = \{ \mathbf{x}_1,...,{\mathbf{x}}_L \} \in \mathbb{R}^{L}$ is mapped to discrete tokens $\hat{\mathbf{H}}_{:n} = \{\hat{\mathbf{h}}_{1}, \dots, \hat{\mathbf{h}}_{N}\} \in \mathbb{R}^{N\times d_{c}}$. Then a downstream forecasting model is trained to learn the mapping between these tokens and the future sequence $\mathbf{Y}=\{ \mathbf{x}_{L+1},...,{\mathbf{x}}_{L+T}\} \in \mathbb{R}^{T}$, performing the time series forecasting task. This process is further elaborated in Section~\ref{predictor}.

\subsection{Predictor}
\label{predictor}
In the second stage, we use the tokenizer, trained in the first stage, to convert a batch of historical sequence $\mathbf{X} \in \mathbb{R}^{B\times L}$ into deeply disentangled discrete tokens $\hat{\mathbf{H}} \in \mathbb{R}^{B \times N \times d_{c}}$. Then these tokens are fed into a downstream forecasting model to obtain the prediction results $\hat{\mathbf{Y}} = \mathcal{F} ( \hat{\mathbf{H}} ) \in \mathbb{R}^{B \times T}$, where $\mathcal{F} ( \cdot )$ represents any suitable forecasting model. 
The tokenizer is designed to be compatible with a wide range of downstream architectures, including Transformer-based models, linear models, and even Large Language Models, offering flexibility in model selection based on the specific needs of different forecasting tasks.

For our experiments, we have selected three widely recognized models: PatchTST~\cite{Yuqietal-2023-PatchTST}, PatchMLP~\cite{Tang2025UnlockingTP}, and GPT4TS~\cite{Zhou2023OneFA} as downstream prediction models.
These models feature highly generalizable architectures and have demonstrated their effectiveness in time series forecasting tasks. We replaced the patch-based tokens in the models with the discrete tokens generated by our method. By combining these diverse prediction models, we are able to evaluate the effectiveness of our approach across different architectures, providing a more comprehensive assessment of its performance.

\section{Experiments}
We conduct extensive experiments to evaluate the effectiveness of \model{} and to validate the contribution of its two core components: \textbf{Deep Fourier Tokenization (DFT)} and the \textbf{Routing Codebook}. Our goal is to answer the following key questions:
\begin{itemize} 
    \item \textbf{RQ1:} Does \model{} improve in-domain prediction?
    \item \textbf{RQ2:} Can \model{} improve the cross-domain generalization ability of traditional supervised models?
    \item \textbf{RQ3:} How do the individual components (DFT and Routing Codebook) of \model{} contribute to its overall performance?
    \item \textbf{RQ4:} Does DFT effectively decouple time series?
    \item \textbf{RQ5:} How does the performance of \model{} change with different hyperparameters?
\end{itemize}

\subsection{Experiment Setup}
\label{appendix:experiment setup}
\textbf{Datasets.}
We extensively include eight commonly used real-world datasets in our experiments, including ETT (four subsets: ETTh1, ETTh2, ETTm1, ETTm2)~\cite{haoyietal-informer-2021}, ECL, Weather, Traffic~\cite{Wu2021AutoformerDT}, and Solar~\cite{lai2018modeling}. Each dataset was divided chronologically with a 6:2:2 ratio for training, validation, and testing. Further details of these datasets can be found in Appendix~\ref{app:datasets}.

\textbf{Baselines.}
\model{} is a flexible two-stage architecture that can seamlessly integrate various downstream forecasting models as its backbone. To demonstrate this versatility, we implement three variants: \modelm{}, \modelt{}, and \modelg{}, using PatchMLP~\cite{Tang2025UnlockingTP}, PatchTST~\cite{Yuqietal-2023-PatchTST}, and GPT4TS~\cite{Zhou2023OneFA} as backbones, respectively. Furthermore, to evaluate the competitive performance of \model{}, we conduct extensive comparisons against a diverse set of state-of-the-art (SOTA) baselines. These include TimeMixer~\cite{wang2023timemixer}, TimesNet~\cite{wu2023timesnet}, FEDformer~\cite{Zhou2022FEDformerFE}, DLinear~\cite{Zeng2022AreTE}, TOTEM~\cite{Talukder2024TOTEMTT}, Moirai~\cite{woo2024moirai}, Chronos~\cite{ansari2024chronos} and Time-MOE~\cite{xiaoming2025time}, covering linear, Transformer-based, VQ-based, and pre-trained forecasting paradigms. Detailed configurations for both our variants and the baselines are provided in Appendix~\ref{app:model_details}.
Unless otherwise stated, we will compare \model{} with the baseline models under the same hyperparameter settings (including input and forecast lengths). 

\textbf{Implementation Details.}
To ensure a fair comparison of model performance across different forecast horizons, we fix the input length at $L = 96$ in all benchmark tests and evaluate models with a wide range of forecast lengths $T = \{96,192,336,720\}$. We choose the Mean Squared Error ($\text{MSE}$) and the Mean Absolute Error ($\text{MAE}$) as metrics to evaluate model performance. Please refer to Appendix~\ref{app:implementation_details} for further details on the implementation. 
The link to the anonymous code repository is provided in the supplementary materials.

\subsection{In-domain Forecast (RQ1)}
\textbf{Setup.}
We evaluate the in-domain prediction performance of three \model{} variants across eight real-world datasets, benchmarking them against a diverse set of baseline models. Table~\ref{tab:in-domain} summarizes the average metrics for all prediction lengths on each dataset. The full experimental results are reported in Appendix~\ref{app:full_results_in_domain}.

\begin{table*}[!ht]
\centering
\caption{Comparison of in-domain forecasting performance. Best results are highlighted in \textbf{bold}, and second best results are \underline{underlined}. Light red shading indicates improvements over the corresponding backbone.}
\label{tab:in-domain}
\resizebox{1\linewidth}{!}{
\begin{tabular}{c||cccccc|cccccc|cc|cccccccc}
\toprule
\multicolumn{1}{c}{\textbf{Method}} & \multicolumn{2}{c}{\textbf{DFTNet(T)}} & \multicolumn{2}{c}{\textbf{DFTNet(G)}} & \multicolumn{2}{c}{\textbf{DFTNet(M)}} & \multicolumn{2}{c}{\textbf{PatchTST}} & \multicolumn{2}{c}{\textbf{GPT4TS}} & \multicolumn{2}{c}{\textbf{PatchMLP}} & \multicolumn{2}{c}{\textbf{TOTEM}} &
\multicolumn{2}{c}{\textbf{TimeMixer}} & \multicolumn{2}{c}{\textbf{TimesNet}} &
\multicolumn{2}{c}{\textbf{DLinear}} & \multicolumn{2}{c}{\textbf{FEDformer}}\\

 \cmidrule(lr){1-1} \cmidrule(lr){2-7} \cmidrule(lr){8-13} \cmidrule(lr){14-15}  \cmidrule(lr){16-23}
\multicolumn{1}{c}{\textbf{Metric}} & \multicolumn{1}{c}{MSE} & \multicolumn{1}{c}{MAE} & \multicolumn{1}{c}{MSE} & \multicolumn{1}{c}{MAE} & \multicolumn{1}{c}{MSE} & \multicolumn{1}{c}{MAE} & \multicolumn{1}{c}{MSE} & \multicolumn{1}{c}{MAE} & \multicolumn{1}{c}{MSE} & \multicolumn{1}{c}{MAE} & \multicolumn{1}{c}{MSE} & \multicolumn{1}{c}{MAE} & \multicolumn{1}{c}{MSE} & \multicolumn{1}{c}{MAE} & \multicolumn{1}{c}{MSE} & \multicolumn{1}{c}{MAE} & \multicolumn{1}{c}{MSE} & \multicolumn{1}{c}{MAE} & \multicolumn{1}{c}{MSE} & \multicolumn{1}{c}{MAE} & \multicolumn{1}{c}{MSE} & \multicolumn{1}{c}{MAE} \\
\midrule

\multirow{1}{*}{\rotatebox{0}{$\text{ETTh1}$}}
& \cellcolor{red!10}\underline{0.538}  & \cellcolor{red!10}\underline{0.509}  & \cellcolor{red!10}0.561  & \cellcolor{red!10}0.520  & 0.557  & \cellcolor{red!10}\textbf{0.508}  & 0.540  & 0.516  & 0.603  & 0.527  & 0.553  & 0.512  & \textbf{0.524}  & 0.518  & 0.573  & 0.523  & 0.578  & 0.529  & 0.547  & \textbf{0.508}  & 0.582  & 0.528 \\

\multirow{1}{*}{\rotatebox{0}{$\text{ETTh2}$}}
& \cellcolor{red!10}0.302  & \cellcolor{red!10}0.365  & \cellcolor{red!10}0.305  & \cellcolor{red!10}0.369  & \cellcolor{red!10}\textbf{0.297}  & \cellcolor{red!10}\textbf{0.361}  & 0.323  & 0.381  & 0.317  & 0.373  & 0.302  & \underline{0.365}  & \underline{0.300}  & 0.375  & 0.317  & 0.373  & 0.324  & 0.378  & 0.316  & 0.380  & 0.362  & 0.407 \\

\multirow{1}{*}{\rotatebox{0}{$\text{ETTm1}$}}
& \cellcolor{red!10}\underline{0.467}  & \cellcolor{red!10}0.439  & \cellcolor{red!10}0.478  & \cellcolor{red!10}0.445  & 0.481  & \cellcolor{red!10}\underline{0.436}  & 0.481  & 0.445  & 0.505  & 0.454  & 0.481  & 0.440  & 0.513  & 0.480  & 0.472  & 0.437  & 0.478  & 0.439  & \textbf{0.463}  & \textbf{0.428}  & 0.600  & 0.520 \\

\multirow{1}{*}{\rotatebox{0}{$\text{ETTm2}$}}
& \cellcolor{red!10}\textbf{0.220}  & \cellcolor{red!10}\textbf{0.303}  & \cellcolor{red!10}0.222  & \cellcolor{red!10}0.306  & \textbf{0.220}  & \cellcolor{red!10}\textbf{0.303}  & \underline{0.221}  & 0.305  & 0.223  & 0.307  & \textbf{0.220}  & \underline{0.304}  & 0.229  & 0.318  & \underline{0.221}  & \underline{0.304}  & 0.222  & \underline{0.304}  & \underline{0.221}  & 0.307  & 0.235  & 0.322 \\

\multirow{1}{*}{\rotatebox{0}{$\text{Weather}$}}
& \cellcolor{red!10}0.272  & 0.285  & \cellcolor{red!10}0.278  & \cellcolor{red!10}0.293  & \cellcolor{red!10}\underline{0.271}  & \cellcolor{red!10}\textbf{0.283}  & 0.273  & 0.285  & 0.286  & 0.295  & 0.272  & \underline{0.284}  & \textbf{0.253}  & 0.302  & 0.273  & 0.288  & 0.278  & 0.294  & 0.275  & 0.298  & 0.287  & 0.314 \\

\multirow{1}{*}{\rotatebox{0}{$\text{ECL}$}}
& 0.201  & \cellcolor{red!10}\underline{0.276}  & 0.212  & 0.285  & \cellcolor{red!10}0.203  & \cellcolor{red!10}0.277  & \textbf{0.198}  & 0.280  & \underline{0.200}  & \textbf{0.273}  & 0.207  & 0.278  & 0.204  & 0.286  & 0.223  & 0.303  & 0.218  & 0.292  & 0.220  & 0.305  & 0.308  & 0.384 \\

\multirow{1}{*}{\rotatebox{0}{$\text{Traffic}$}}
& \underline{0.487}  & 0.286  & 0.500  & 0.292  & 0.526  & 0.319  & \textbf{0.470}  & \textbf{0.267}  & 0.499  & \underline{0.279}  & 0.493  & 0.285  & 0.579  & 0.294  & 0.611  & 0.349  & 0.538  & 0.310  & 0.657  & 0.366  & 0.887  & 0.467 \\

\multirow{1}{*}{\rotatebox{0}{$\text{Solar}$}}
& \cellcolor{red!10}\underline{0.245}  & \cellcolor{red!10}\underline{0.249}  & 0.289  & 0.268  & \cellcolor{red!10}0.267  & \cellcolor{red!10}\textbf{0.247}  & 0.256  & 0.267  & 0.271  & 0.253  & 0.269  & 0.255  & \textbf{0.229}  & 0.291  & 0.316  & 0.294  & 0.354  & 0.316  & 0.327  & 0.307  & 0.337  & 0.383 \\

\bottomrule
\end{tabular}
}
\end{table*}

\textbf{Analysis.}
As illustrated in Table~\ref{tab:in-domain}, \model{} consistently enhances the performance of all three baseline backbones across diverse datasets, with \modelm{} achieving the most superior results. Furthermore, \model{} variants exhibit strong competitiveness when compared against the VQ-based TOTEM and a broad range of state-of-the-art (SOTA) models.
These findings demonstrate that \model{} provides direct in-domain forecasting gains for models with varying architectures. Although the VQ-based quantization process may introduce minor information loss, the deep basis derived from our deep decoupling mechanism yields a more informative representation than traditional patch-based tokenization. This enriched representation effectively compensates for quantization errors and underpins the model's improved predictive performance.

\subsection{Cross-domain Forecast (One to One) (RQ2)}
\label{cross_domain_one_to_one}
\textbf{Setup.}
To evaluate the cross-domain generalizability of each model, we first adopt a one-to-one setting. Specifically, the model is trained on one dataset (source domain) and directly evaluated on a different dataset (target domain) without any fine-tuning. 
We use four subsets of the ETT to construct six zero-shot generalization tasks. The full experimental results are reported in Appendix~\ref{app:full_results_one_to_one}. 

\textbf{Analysis.}
Table~\ref{tab:cross-domain-one-to-one} reports the average error metrics across all forecasting horizons. The \model{} variants consistently outperform their respective baseline backbones in the majority of cross-domain tasks, demonstrating remarkable robustness in zero-shot scenarios and surpassing most baseline models. These results suggest that the structured deep basis extracted by \model{} possesses superior transferability across diverse domains, thereby effectively bolstering the generalization capabilities of the underlying backbone models.

\begin{table*}[!ht]
\centering
\caption{Comparison of cross-domain (one to one) forecasting performance. Best results are highlighted in \textbf{bold}, and second best results are \underline{underlined}. Light red shading indicates improvements over the corresponding backbone.}
\label{tab:cross-domain-one-to-one}
\resizebox{1\linewidth}{!}{
\begin{tabular}{c||cccccc|cccccc|cc|cccccccc}
\toprule
\multicolumn{1}{c}{\textbf{Method}} & \multicolumn{2}{c}{\textbf{DFTNet(T)}} & \multicolumn{2}{c}{\textbf{DFTNet(G)}} & \multicolumn{2}{c}{\textbf{DFTNet(M)}} & \multicolumn{2}{c}{\textbf{PatchTST}} & \multicolumn{2}{c}{\textbf{GPT4TS}} & \multicolumn{2}{c}{\textbf{PatchMLP}} & \multicolumn{2}{c}{\textbf{TOTEM}} &
\multicolumn{2}{c}{\textbf{TimeMixer}} & \multicolumn{2}{c}{\textbf{TimesNet}} &
\multicolumn{2}{c}{\textbf{DLinear}} & \multicolumn{2}{c}{\textbf{FEDformer}}\\
\cmidrule(lr){1-1} \cmidrule(lr){2-7} \cmidrule(lr){8-13} \cmidrule(lr){14-15}  \cmidrule(lr){16-23}
\multicolumn{1}{c}{\textbf{Metric}} & \multicolumn{1}{c}{MSE} & \multicolumn{1}{c}{MAE} & \multicolumn{1}{c}{MSE} & \multicolumn{1}{c}{MAE} & \multicolumn{1}{c}{MSE} & \multicolumn{1}{c}{MAE} & \multicolumn{1}{c}{MSE} & \multicolumn{1}{c}{MAE} & \multicolumn{1}{c}{MSE} & \multicolumn{1}{c}{MAE} & \multicolumn{1}{c}{MSE} & \multicolumn{1}{c}{MAE} & \multicolumn{1}{c}{MSE} & \multicolumn{1}{c}{MAE} & \multicolumn{1}{c}{MSE} & \multicolumn{1}{c}{MAE} & \multicolumn{1}{c}{MSE} & \multicolumn{1}{c}{MAE} & \multicolumn{1}{c}{MSE} & \multicolumn{1}{c}{MAE} & \multicolumn{1}{c}{MSE} & \multicolumn{1}{c}{MAE} \\
\midrule

\multirow{1}{*}{\rotatebox{0}{$\text{ETTh1}$} $\rightarrow$ \rotatebox{0}{$\text{ETTh2}$}}
& \cellcolor{red!10}\underline{0.320} & \cellcolor{red!10}0.375 & \cellcolor{red!10}0.321 & \cellcolor{red!10}0.375 & \textbf{0.318} & \cellcolor{red!10}\textbf{0.373} & 0.334 & 0.386 & 0.335 & 0.382 & \textbf{0.318} & \underline{0.374} & 0.359 & 0.415 & 0.483 & 0.412 & 0.530 & 0.446 & 0.483 & 0.443 & 0.560 & 0.485 \\

\multirow{1}{*}{\rotatebox{0}{$\text{ETTh1}$} $\rightarrow$ \rotatebox{0}{$\text{ETTm2}$}}
& \cellcolor{red!10}\underline{0.256} & \cellcolor{red!10}\underline{0.345} & \cellcolor{red!10}\textbf{0.253} & \cellcolor{red!10}\textbf{0.343} & \cellcolor{red!10}0.257 & \cellcolor{red!10}0.346 & 0.275 & 0.359 & 0.268 & 0.351 & 0.271 & 0.356 & 0.359 & 0.411 & 0.348 & 0.364 & 0.355 & 0.362 & 0.350 & 0.386 & 0.372 & 0.386 \\

\multirow{1}{*}{\rotatebox{0}{$\text{ETTh2}$} $\rightarrow$ \rotatebox{0}{$\text{ETTh1}$}}
& \cellcolor{red!10}0.644 & \cellcolor{red!10}\underline{0.560} & \cellcolor{red!10}0.664 & 0.581 & \underline{0.629} & \cellcolor{red!10}\textbf{0.550} & 0.798 & 0.635 & 0.682 & 0.574 & \textbf{0.628} & 0.562 & 0.920 & 0.677 & 0.718 & 0.578 & 0.842 & 0.638 & 0.741 & 0.600 & 0.998 & 0.711 \\

\multirow{1}{*}{\rotatebox{0}{$\text{ETTm1}$} $\rightarrow$ \rotatebox{0}{$\text{ETTh2}$}}
& 0.377 & 0.413 & \cellcolor{red!10}0.377 & 0.418 & 0.376 & 0.410 & \underline{0.367} & \underline{0.408} & 0.384 & 0.418 & 0.372 & 0.410 & \textbf{0.338} & \textbf{0.406} & 0.565 & 0.462 & 0.635 & 0.506 & 0.536 & 0.480 & 0.720 & 0.593 \\

\multirow{1}{*}{\rotatebox{0}{$\text{ETTm1}$} $\rightarrow$ \rotatebox{0}{$\text{ETTm2}$}}
& \cellcolor{red!10}\textbf{0.233} & \cellcolor{red!10}\textbf{0.313} & \cellcolor{red!10}\underline{0.234} & \cellcolor{red!10}\underline{0.314} & 0.236 & \cellcolor{red!10}\underline{0.314} & \underline{0.234} & \underline{0.314} & 0.238 & 0.317 & 0.235 & 0.315 & 0.242 & 0.336 & 0.325 & 0.326 & 0.375 & 0.366 & 0.372 & 0.402 & 0.415 & 0.431 \\

\multirow{1}{*}{\rotatebox{0}{$\text{ETTm2}$} $\rightarrow$ \rotatebox{0}{$\text{ETTm1}$}}
& 0.604 & 0.513 & \cellcolor{red!10}\underline{0.572} & \cellcolor{red!10}\underline{0.499} & \cellcolor{red!10}0.588 & \cellcolor{red!10}0.500 & 0.585 & 0.504 & 0.609 & 0.519 & 0.599 & 0.512 & 1.026 & 0.640 & \textbf{0.543} & \textbf{0.488} & 0.857 & 0.626 & 0.688 & 0.555 & 0.793 & 0.631 \\

\bottomrule
\end{tabular}
}
\end{table*}

\subsection{Cross-domain Forecast (Multi to One) (RQ2)}
\textbf{Setup.}
To further evaluate the domain-agnostic nature and scalability of the proposed deep basis, we conduct extensive training across multiple datasets and assess the model's zero-shot generalization capability on an unseen target domain without any fine-tuning. We utilize five diverse datasets: ECL, Weather, Traffic, Solar, and ETT (where the four ETT subsets are treated as a unified domain). In this multi-to-one setting, the model is trained on four datasets and directly evaluated on the remaining one. Given that this paradigm aligns with the development of time-series foundation models, we incorporate a suite of prominent foundation models as additional baselines for a comprehensive comparison. The full experimental results are provided in Appendix~\ref{app:full_results_multi_to_one}.

\textbf{Analysis.}
As presented in Table~\ref{tab:cross-domain-multi-to-one}, \model{} consistently enhances the zero-shot capabilities of its respective backbones across most metrics. It is worth noting that even when trained on a limited set of four datasets, the \model{} variants deliver performance comparable to several foundation models pre-trained on much larger time series repositories. These results further validate the effectiveness of our proposed deep decoupling mechanism, suggesting that the structured deep basis can efficiently capture essential temporal characteristics, thereby providing a promising alternative for robust cross-domain forecasting.

\begin{table*}[!ht]
\centering
\caption{Comparison of cross-domain (multi to one) forecasting performance. Best results are highlighted in \textbf{bold}, and second best results are \underline{underlined}. '-' denotes baselines pre-trained on the target domain and thus excluded from zero-shot comparison. Light red shading indicates improvements over the corresponding backbone.}
\label{tab:cross-domain-multi-to-one}
\resizebox{1\linewidth}{!}{
\begin{tabular}{c||cccccc|cccccc|cc|cccccc}
\toprule
\multicolumn{1}{c}{\textbf{Method}} & \multicolumn{2}{c}{\textbf{DFTNet(T)}} & \multicolumn{2}{c}{\textbf{DFTNet(G)}} & \multicolumn{2}{c}{\textbf{DFTNet(M)}} & \multicolumn{2}{c}{\textbf{PatchTST}} & \multicolumn{2}{c}{\textbf{GPT4TS}} & \multicolumn{2}{c}{\textbf{PatchMLP}} & \multicolumn{2}{c}{\textbf{TOTEM}} &
\multicolumn{2}{c}{\(\textbf{Moirai}_{base}\)} & \multicolumn{2}{c}{\(\textbf{Chronos}_{base}\)} &
\multicolumn{2}{c}{\(\textbf{Time-MOE}_{base}\)} \\

\cmidrule(lr){1-1} \cmidrule(lr){2-7} \cmidrule(lr){8-13} \cmidrule(lr){14-15}  \cmidrule(lr){16-21}

\multicolumn{1}{c}{\textbf{Metric}} & \multicolumn{1}{c}{MSE} & \multicolumn{1}{c}{MAE} & \multicolumn{1}{c}{MSE} & \multicolumn{1}{c}{MAE} & \multicolumn{1}{c}{MSE} & \multicolumn{1}{c}{MAE} & \multicolumn{1}{c}{MSE} & \multicolumn{1}{c}{MAE} & \multicolumn{1}{c}{MSE} & \multicolumn{1}{c}{MAE} & \multicolumn{1}{c}{MSE} & \multicolumn{1}{c}{MAE} & \multicolumn{1}{c}{MSE} & \multicolumn{1}{c}{MAE} & \multicolumn{1}{c}{MSE} & \multicolumn{1}{c}{MAE} & \multicolumn{1}{c}{MSE} & \multicolumn{1}{c}{MAE} & \multicolumn{1}{c}{MSE} & \multicolumn{1}{c}{MAE} \\
\midrule

\multirow{1}{*}{\rotatebox{0}{$\text{Multi}$} $\rightarrow$ \rotatebox{0}{$\text{Traffic}$}}
& \cellcolor{red!10}\textbf{0.707}  & \cellcolor{red!10}\underline{0.404}  & 0.871  & 0.513  & \cellcolor{red!10}\underline{0.736}  & \cellcolor{red!10}0.415  & 0.741  & 0.426  & 0.795  & 0.461  & 0.740  & 0.420  & 0.829  & 0.484  &  -  &  -  & 0.759  & \textbf{0.335}  &  -  &  - \\

\multirow{1}{*}{\rotatebox{0}{$\text{Multi}$} $\rightarrow$ \rotatebox{0}{$\text{Weather}$}}
& \cellcolor{red!10}0.307  & \cellcolor{red!10}\underline{0.313}  & \cellcolor{red!10}\underline{0.303}  & \cellcolor{red!10}0.314  & 0.338  & \cellcolor{red!10}0.316  & 0.308  & 0.316  & 0.316  & 0.317  & 0.334  & 0.317  & 0.333  & 0.378  & \textbf{0.284}  & \textbf{0.292}  & 0.431  & 0.336  & 0.404  & 0.402 \\
 
\multirow{1}{*}{\rotatebox{0}{$\text{Multi}$} $\rightarrow$ \rotatebox{0}{$\text{ECL}$}}
& \cellcolor{red!10}0.269  & \cellcolor{red!10}0.340  & \cellcolor{red!10}0.308  & \cellcolor{red!10}0.382  & 0.298  & \cellcolor{red!10}0.349  & 0.301  & 0.361  & 0.312  & 0.383  & 0.298  & 0.362  & 0.291  & 0.354  & \underline{0.251}  & \textbf{0.314}  &  -  &  -  & \textbf{0.245}  & \underline{0.327} \\


 
\bottomrule
\end{tabular}
}
\end{table*}

\begin{figure*}[!ht]
    \centering
    \begin{minipage}[b]{0.98\textwidth} 
        \centering
        \begin{subfigure}[b]{0.31\textwidth}
            \centering
            \includegraphics[width=\textwidth]{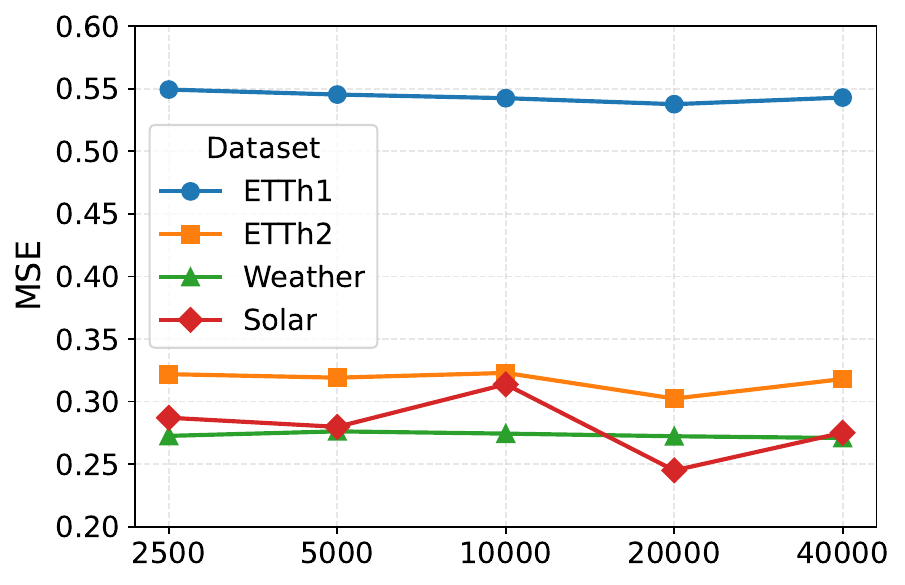}
            \caption{Codebook Size \(K\)}
            \label{fig:hyperparameter_study_codebook_size}
        \end{subfigure}
        \hfill
        \begin{subfigure}[b]{0.31\textwidth}
            \centering
            \includegraphics[width=\textwidth]{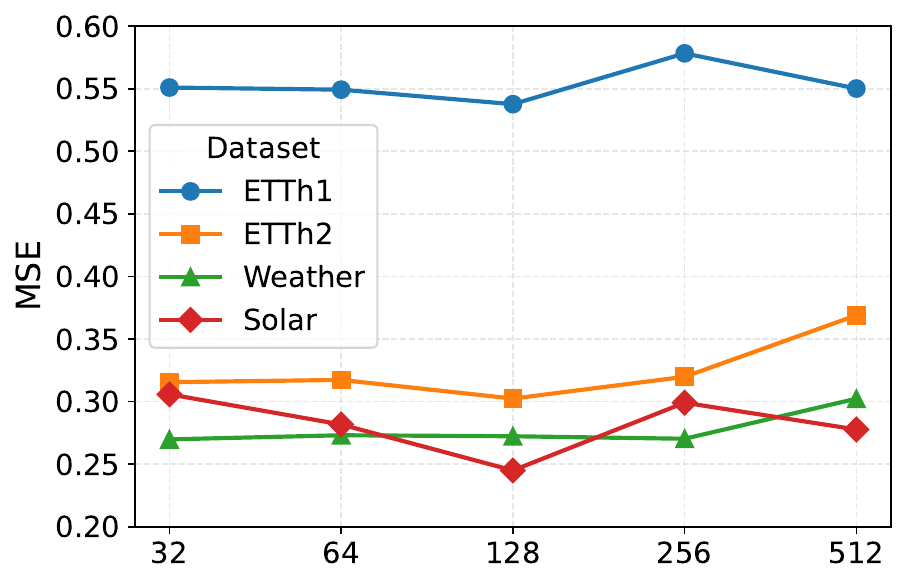}
            \caption{Code Dim \(d_c\)}
            \label{fig:hyperparameter_study_code_dim}
        \end{subfigure}
        \hfill
        \begin{subfigure}[b]{0.35\textwidth}
            \centering
            \includegraphics[width=\textwidth]{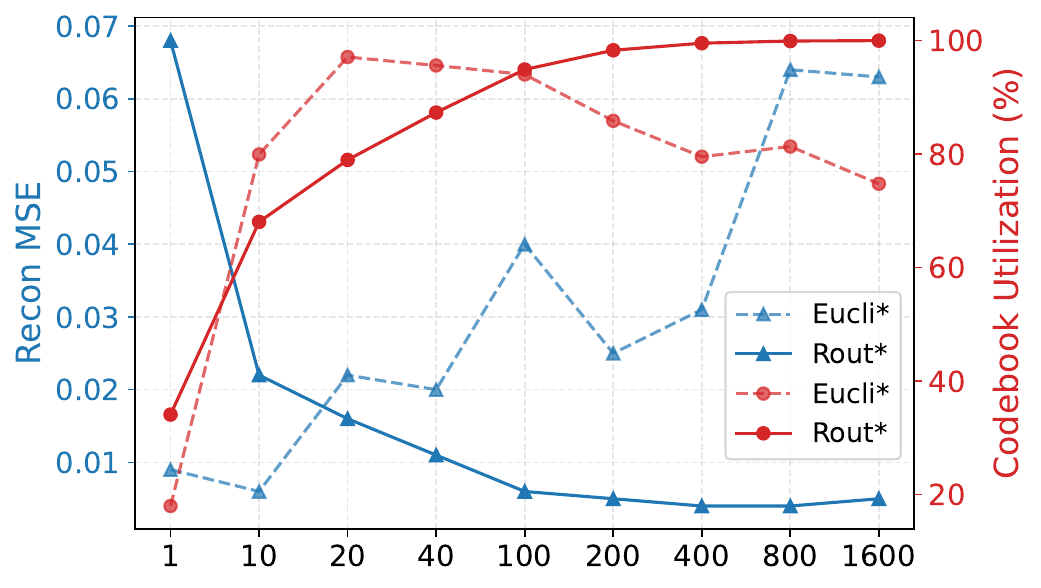}
            \caption{Selected Codewords Number \(\text{top-}k\)}
            \label{fig:hyperparameter_study_top_k}
        \end{subfigure}
    \end{minipage}
    \caption{Hyperparameter Study of \model{}}
    \label{fig:hyperparameter_study_all}
    
\end{figure*}

\subsection{Ablation Study (RQ3)}
\label{Ablation Study}
In this section, to assess the effectiveness of the core components of \model{}, we conduct ablation studies on ETTh1, ETTh2, Weather, and Solar. Specifically, we analyze five model variants with individual components removed: 
\textbf{(1) w/o $\text{DFT}(\mathcal{L}_{\text{ortho}})$}: remove the orthogonality loss from DFT; 
\textbf{(2) w/o $\text{DFT}(\mathcal{L}_{\text{consis}})$}: remove the consistency loss from DFT; 
\textbf{(3) w/o $\text{DFT(all)}$}: remove the entire DFT; 
\textbf{(4) w/o $\text{Routing}$}: remove the routing mechanism in Routing Codebook and use the original Euclidean distance for codeword selection; 
\textbf{(5) w/o $\text{DFT(all) and Routing}$}: remove both DFT and Routing Codebook. 

Table~\ref{table:ablation} presents the average performance of \model{} and its five variants on the ETTh1 dataset across all prediction lengths, using $\text{MSE}$ and $\text{MAE}$ as evaluation metrics. The complete removal of DFT (\textbf{w/o DFT}) directly impairs performance, underscoring its critical role in time series decomposition for prediction. The model performance degrades further when either the orthogonality loss (\textbf{w/o $\text{DFT}(\mathcal{L}_{\text{ortho}})$}) or the consistency loss (\textbf{w/o $\text{DFT}(\mathcal{L}_{\text{consis}})$}) is individually removed, which signifies the necessity of their simultaneous optimization. Moreover, eliminating the Routing Codebook (\textbf{w/o Routing}) diminishes performance, confirming the superiority of our learnable routing mechanism over traditional Euclidean distance. Finally, the variant lacking both core components (\textbf{w/o DFT(all) and Routing}) exhibits the worst performance, reaffirming their importance to \model{}. 



\begin{table}[h]
\centering
\caption{Average MSE and MAE of the ablated variants of \model{} on multiple datasets.}
\label{table:ablation}
\resizebox{0.8\linewidth}{!}{
\begin{tabular}{l cc cc cc cc}
\toprule
\multicolumn{1}{c}{\textbf{Dataset}} & \multicolumn{2}{c}{\textbf{ETTh1}} & \multicolumn{2}{c}{\textbf{ETTh2}} & \multicolumn{2}{c}{\textbf{Weather}} & \multicolumn{2}{c}{\textbf{Solar}} \\
\cmidrule(lr){1-1} \cmidrule(lr){2-3} \cmidrule(lr){4-5} \cmidrule(lr){6-7}  \cmidrule(lr){8-9}
\multicolumn{1}{c}{\textbf{Metric}} & \multicolumn{1}{c}{MSE} & \multicolumn{1}{c}{MAE} & \multicolumn{1}{c}{MSE} & \multicolumn{1}{c}{MAE} & \multicolumn{1}{c}{MSE} & \multicolumn{1}{c}{MAE} & \multicolumn{1}{c}{MSE} & \multicolumn{1}{c}{MAE} \\
\midrule
\model{} & \textbf{0.5376} & \textbf{0.5091} & \textbf{0.3024} & \textbf{0.3650} & \textbf{0.2723} & \textbf{0.2846} & \textbf{0.2450} & \textbf{0.2490} \\
w/o $\text{DFT}(\mathcal{L}_{\text{ortho}})$ & 0.5479 & \underline{0.5113} & 0.3190 & \underline{0.3746} & \underline{0.2739} & \underline{0.2910} & 0.2839 & 0.2783 \\
w/o $\text{DFT}(\mathcal{L}_{\text{consis}})$ & 0.5505 & 0.5200 & 0.3179 & 0.3759 & 0.2770 & 0.2930 & 0.2735 & 0.2697 \\
w/o $\text{DFT(all)}$ & \underline{0.5436} & 0.5154 & \underline{0.3175} & 0.3750 & 0.2768 & 0.2926 & \underline{0.2732} & \underline{0.2685} \\
w/o $\text{Routing}$ & 0.5503 & 0.5164 & 0.3217 & 0.3782 & 0.2794 & 0.2962 & 0.2835 & 0.2806 \\ 
w/o $\text{DFT(all)~and~Routing}$ & 0.5595 & 0.5200 & 0.3196 & 0.3751 & 0.2801 & 0.2962 & 0.2844 & 0.2794 \\
\bottomrule   
\end{tabular}
}
\end{table}

\subsection{Visualization (RQ4)}
To evaluate the effectiveness of \model{} in decoupling time series representations, we employed UMAP~\cite{mcinnes2018umap-software} for dimensionality reduction and visualization of the time series bases ($\mathbf{H}^{DFT} \in \mathbb{R}^{B \times N \times d_{c}}$) extracted from the trained DFT module. 
Appendix~\ref{app:visualization} provides additional visualization results on various datasets in Figure~\ref{fig:visualization_umap}.
As illustrated, bases from different categories are well separated, whereas those within the same category form tight clusters. This clearly demonstrates the efficacy of the orthogonality loss $\mathcal{L}_{\text{ortho}}$ and the consistency loss $\mathcal{L}_{\text{consis}}$ introduced in our model to achieve effective decomposition of time series.

\subsection{Hyper-parameters Study (RQ5)}
This section analyzes the influence of hyperparameters on model performance. Specifically, using $\text{MSE}$ as the evaluation metric, we mainly investigate three critical hyperparameters in \model{}: \(K\) (codebook size), \(d_c\) (code dim), and \(\text{top-}k\) (selected codewords number). 
In addition, we report a study of the hyperparameters of loss weight \(\lambda / \alpha\) in Appendix~\ref{app:additional_ablation_study_loss_weight}. 

\textbf{Codebook Size \(K\). }
The codebook size determines the number of codewords. As shown in Figure~\ref{fig:hyperparameter_study_codebook_size}, model performance improves with \(K\), but not monotonically. While a larger \(K\) enhances discrete representations and captures more features, excessively large \(K\) reduces the codebook's ability to compress information and minimize redundancy, ultimately leading to worse performance.

\textbf{Code Dim \(d_c\). }
The code dim specifies the dimensionality of each codeword. A smaller \(d_c\) implies simpler information encoding, while a larger \(d_c\) can represent more complex features but risks overfitting. As show in Figure~\ref{fig:hyperparameter_study_code_dim}, the model achieves optimal performance when \(d_c\) is 128. Performance deteriorates to varying degrees across all prediction lengths when \(d_c\) is either too large or too small.

\textbf{Selected Codewords Number} \(\text{top-}k\)\textbf{. }
The \(\text{top-}k\) parameter plays a crucial role in the routing mechanism, impacting both reconstruction performance and codebook utilization. As shown in Figure~\ref{fig:hyperparameter_study_top_k}, the Routing Codebook (Rout*) shows continuous improvement in reconstruction loss (MSE) and codebook utilization as \(\text{top-}k\) increases. In contrast, the Euclidean Codebook (Eucl*) does not consistently benefit from larger 
\(\text{top-}k\), with both reconstruction performance and codebook utilization declining.

\subsection{Efficiency Analysis}
We perform an efficiency analysis using \modelt{} as a representative model and compare its performance with the baseline. Detailed results are provided in Appendix~\ref{app:efficiency_analysis}. 

\section{Conclusion}
This paper aims to enhance the unified modeling capability across multi-domain time series and achieve competitive generalization by proposing a novel vector quantization-based time series tokenization method. We introduce an innovative time series forecasting model (\model{}). By incorporating Deep Fourier Tokenization (DFT) and learnable Routing Codebook, we effectively address the limitations of existing methods in time series decoupling and codebook utilization. 
As a universal discrete tokenization approach, our method achieves significant performance improvements across eight real-world time series datasets. 
Extensive in-domain and cross-domain forecasting experiments demonstrate the effectiveness of the structured discrete representations provided by \model{} in enhancing the model's universal forecasting and cross-domain generalization capabilities.

\bibliographystyle{unsrtnat}
\bibliography{references}
\newpage
\appendix

\section{Datasets}
\label{app:datasets}
We evaluate our model on eight publicly available multivariate time series datasets, covering a wide range of domains and temporal characteristics. The ETTh1 and ETTh2 datasets consist of hourly measurements of electricity transformer temperature from two different locations, while ETTm1 and ETTm2 are their high-resolution counterparts sampled every 15 minutes. The Electricity Consumption Load (ECL) dataset contains hourly usage data from 321 clients, capturing diverse consumption behaviors. The Weather dataset includes meteorological variables (e.g., temperature, humidity, wind speed) from 21 sensors sampled every 10 minutes.

In addition, the Traffic dataset records the hourly occupancy rates of 862 sensors on San Francisco Bay Area highways, exhibiting strong periodic patterns and high dimensionality. The Solar dataset provides solar power output from 137 stations at 10-minute intervals, characterized by clear daily cycles and sparse nighttime activity. A summary of the dataset statistics is provided in Table~\ref{tab:dataset}.


\begin{table}[h]
\centering
\caption{Dataset statistics.}
\small
\begin{tabular}{@{}lcccc@{}}
\toprule
Datasets & Channels & Time Steps & Frequency          \\ 
\midrule
ETTh1, ETTh2    & 7        & 17472    & hourly \\
ETTm1, ETTm2    & 7        & 69680    & 15-min \\
ECL      & 321      & 26304    & hourly \\
Weather  & 21       & 52696    & 10-min \\
Traffic  & 862      & 17544    & hourly \\
Solar    & 137      & 52560    & 10-min \\ 
\bottomrule
\end{tabular}
\label{tab:dataset}
\end{table}

\section{Model Details}
\label{app:model_details}

\subsection{Baselines}
\label{app:baseline_details}
To ensure a comprehensive evaluation across various prediction tasks, we benchmark \model{} against a diverse set of baseline models, categorized into four groups based on their architectural characteristics and our evaluation objectives:

\textbf{Backbone Architectures. }
We employ PatchTST~\cite{Yuqietal-2023-PatchTST}, GPT4TS~\cite{Zhou2023OneFA}, and PatchMLP~\cite{Tang2025UnlockingTP} as the foundational backbones to implement three \model{} variants. Comparing our variants directly against these base models allows us to isolate and quantify the performance gains specifically attributed to the \model{} framework.

\textbf{State-of-the-Art Supervised Models. }
To evaluate in-domain and near-domain (cross-domain (one to one)) zero-shot performance, we include a suite of widely recognized supervised models, such as TimeMixer~\cite{wang2023timemixer}, TimesNet~\cite{wu2023timesnet}, DLinear~\cite{Zeng2022AreTE}, and FEDformer~\cite{Zhou2022FEDformerFE}. This selection encompasses both Transformer-based and linear paradigms, providing a robust benchmark to situate \model{} within the current landscape of supervised forecasting.

\textbf{Foundation Models. }
For cross-domain (multi-to-one) scenarios—where models are trained on multiple domains and evaluated on an unseen target domain—we incorporate prominent zero-shot learners, including Moirai~\cite{woo2024moirai}, Chronos~\cite{ansari2024chronos}, and Time-MOE~\cite{xiaoming2025time}. To ensure a fair comparison, we utilize the 'base' versions of these models to assess the relative generalization capabilities of \model{}.

\textbf{VQ-based Paradigms. }
Given that \model{} is built upon the Vector Quantization (VQ) framework, we additionally introduce TOTEM~\cite{Talukder2024TOTEMTT}, a high-performance VQ-based model, as a specialized baseline. This comparison serves to highlight the architectural advancements and performance enhancements introduced by \model{} over existing VQ-based forecasting methods.

\subsection{DFTNet Variants}
\label{app:dft_variants_details}
This section elaborates on the implementation details of \model{} incorporating various downstream prediction models, including \modelt{}, \modelg{}, and \modelm{}. Given that \model{} serves as a general time series tokenization method, when enhancing each baseline model, we only replace its tokenization mechanism while keeping other parts of the model unchanged. We provide complete, reproducible code in the anonymous repository, and the link to the repository is provided in the supplementary materials.

\textbf{Routing Codebook. }
In the Routing Codebook, the Encoder and Decoder used are simple linear layer structures; the activation function $\sigma(\cdot)$ in the Routing MLP is the $\text{LeakyReLU}(\cdot)$ activation function; and the $\beta$ value of the commitment loss and the EMA update rate of the codebook are set to 0.1 and 0.2, respectively.

\textbf{\modelt{}. }
\modelt{} employs PatchTST~\cite{Yuqietal-2023-PatchTST} as its downstream prediction model. We utilize the time series tokenizer trained in the first stage (Section~\ref{stage_1}) to replace the original $\text{patch embedding}(\cdot)$ operation applied to input sequences in PatchTST, thereby generating discrete representations of the sequences.

\textbf{\modelg{}. }
\modelg{} is built upon GPT4TS~\cite{Zhou2023OneFA}. Following the recommendations from the original article, we set the gpt layer parameter to 6. Similarly, we replace the patch operation within the model with the tokenizer trained in the first stage. As our chosen $\text{GPT2}$ backbone receives fixed 768-dimensional representations as input, during the first stage of training, we uniformly adjusted both the dimension of the DFT decoupled basis and the dimension of the Codewords to 768 to ensure compatibility with the downstream prediction model. 

\textbf{\modelm{}. }
PatchMLP~\cite{Tang2025UnlockingTP} adopts a multi-scale patch mixing strategy. \modelm{} directly replaces the corresponding component of PatchMLP with the tokenizer trained in the first stage, while retaining its core model architecture. 

\section{Implementation Details}
\label{app:implementation_details}
\textbf{Evaluation Metrics.}
To comprehensively evaluate the model's performance, we utilized two common forecasting evaluation metrics: Mean Square Error ($\text{MSE}$) and Mean Absolute Error ($\text{MAE}$).

$\text{MAE}$ measures the average absolute difference between predicted values and actual values. The formula for $\text{MAE}$ is defined as follows:
\begin{equation}
    \mathrm{MAE}=\frac{1}{T}\sum_{\mathrm{i=1}}^T|\mathbf{Y}_i-\hat{\mathbf{Y}}_i|
\end{equation}
Here, $\mathbf{Y}_i$ represents the actual value, $\hat{\mathbf{Y}}_i$ represents the predicted value, and $T$ denotes the number of future timestamps to be predicted, the same applies to the following equations. $\text{MAE}$ retains the same unit as the target variable, making it intuitively interpretable. Compared to squared-error metrics, $\text{MAE}$ is less sensitive to outliers and provides a robust estimate of the average forecasting deviation.

$\text{MSE}$ quantifies the average squared difference between predicted and actual values, calculated as:
\begin{equation}
    \mathrm{MSE}=\frac{1}{T}\sum_{\mathrm{i=1}}^T(\mathbf{Y}_i-\hat{\mathbf{Y}}_i)^2
\end{equation}
$\text{MSE}$ penalizes larger errors more heavily due to the squared term, making it particularly sensitive to outliers. This sensitivity helps highlight models that occasionally make large errors, providing a stricter evaluation of forecasting performance in such cases.

\textbf{Training Details.}
During training, all models are optimized using the NAdam optimizer with an initial learning rate of 0.001. We employ a type1 learning rate scheduler and set the maximum number of training epochs to 100, with an early stopping patience of five epochs. In the first stage, $\text{MSE}$ was used as the loss function and $\text{MAE}$ in the second stage. The batch size is fixed at 512 across all datasets. Each experiment is repeated five times, with metric average results reported. All experiments are implemented in PyTorch using two NVIDIA H100 80GB HBM3 GPUs.

\section{Full Experimental Results}
\label{app:full_results}
This section aims to present in detail the experimental results that could not be fully displayed in the main text due to space constraints.

\begin{table*}[h]
\centering
\caption{Full results of in-domain forecasting performance. Best results are highlighted in \textbf{bold}, and second best results are \underline{underlined}. Light red shading indicates improvements over the corresponding backbone.}
\label{tab:appendix:in_domain}
\begin{center}
\resizebox{1\linewidth}{!}{
\begin{tabular}{cc||cccccc|cccccc|cc|cccccccc}
\toprule
\multicolumn{2}{c}{\textbf{Method}} & \multicolumn{2}{c}{\textbf{DFTNet(T)}} & \multicolumn{2}{c}{\textbf{DFTNet(G)}} & \multicolumn{2}{c}{\textbf{DFTNet(M)}} & \multicolumn{2}{c}{\textbf{PatchTST}} & \multicolumn{2}{c}{\textbf{GPT4TS}} & \multicolumn{2}{c}{\textbf{PatchMLP}} & \multicolumn{2}{c}{\textbf{TOTEM}} &
\multicolumn{2}{c}{\textbf{TimeMixer}} & \multicolumn{2}{c}{\textbf{TimesNet}} &
\multicolumn{2}{c}{\textbf{DLinear}} & \multicolumn{2}{c}{\textbf{FEDformer}}\\

 \cmidrule(lr){1-2} \cmidrule(lr){3-8} \cmidrule(lr){9-14} \cmidrule(lr){15-16}  \cmidrule(lr){17-24}
\multicolumn{2}{c}{\textbf{Metric}} & \multicolumn{1}{c}{MSE} & \multicolumn{1}{c}{MAE} & \multicolumn{1}{c}{MSE} & \multicolumn{1}{c}{MAE} & \multicolumn{1}{c}{MSE} & \multicolumn{1}{c}{MAE} & \multicolumn{1}{c}{MSE} & \multicolumn{1}{c}{MAE} & \multicolumn{1}{c}{MSE} & \multicolumn{1}{c}{MAE} & \multicolumn{1}{c}{MSE} & \multicolumn{1}{c}{MAE} & \multicolumn{1}{c}{MSE} & \multicolumn{1}{c}{MAE} & \multicolumn{1}{c}{MSE} & \multicolumn{1}{c}{MAE} & \multicolumn{1}{c}{MSE} & \multicolumn{1}{c}{MAE} & \multicolumn{1}{c}{MSE} & \multicolumn{1}{c}{MAE} & \multicolumn{1}{c}{MSE} & \multicolumn{1}{c}{MAE} \\
\midrule

\multirow{5}{*}{\rotatebox{0}{$\text{ETTh1}$}}
& 96  & 0.433  & \cellcolor{red!10}\underline{0.440}  & \cellcolor{red!10}0.459  & \cellcolor{red!10}0.453  & \cellcolor{red!10}0.456  & \cellcolor{red!10}\textbf{0.439}  & \textbf{0.424}  & 0.441  & 0.499  & 0.458  & 0.464  & 0.448  & \underline{0.432}  & 0.454  & 0.475  & 0.454  & 0.478  & 0.459  & 0.468  & 0.450  & 0.471  & 0.460 \\
& 192 & \cellcolor{red!10}\underline{0.492}  & \cellcolor{red!10}0.479  & \cellcolor{red!10}0.517  & \cellcolor{red!10}0.494  & 0.511  & \underline{0.478}  & 0.496  & 0.487  & 0.547  & 0.495  & 0.493  & \textbf{0.476}  & \textbf{0.485}  & 0.491  & 0.530  & 0.494  & 0.532  & 0.497  & 0.518  & 0.484  & 0.530  & 0.498 \\
& 336 & \underline{0.542}  & \cellcolor{red!10}\underline{0.514}  & \cellcolor{red!10}0.567  & \cellcolor{red!10}0.528  & 0.554  & \cellcolor{red!10}\textbf{0.511}  & \underline{0.542}  & 0.520  & 0.597  & 0.530  & 0.546  & \underline{0.514}  & \textbf{0.533}  & 0.530  & 0.582  & 0.532  & 0.583  & 0.538  & 0.553  & \underline{0.514}  & 0.576  & 0.529 \\
& 720 & \cellcolor{red!10}0.684  & \cellcolor{red!10}0.603  & \cellcolor{red!10}0.701  & \cellcolor{red!10}0.606  & \cellcolor{red!10}0.707  & \cellcolor{red!10}0.603  & 0.699  & 0.615  & 0.771  & 0.627  & 0.708  & 0.610  & \textbf{0.647}  & \underline{0.599}  & 0.707  & 0.612  & 0.717  & 0.621  & \underline{0.649}  & \textbf{0.584}  & 0.749  & 0.626 \\
& Avg & \cellcolor{red!10}\underline{0.538}  & \cellcolor{red!10}\underline{0.509}  & \cellcolor{red!10}0.561  & \cellcolor{red!10}0.520  & 0.557  & \cellcolor{red!10}\textbf{0.508}  & 0.540  & 0.516  & 0.603  & 0.527  & 0.553  & 0.512  & \textbf{0.524}  & 0.518  & 0.573  & 0.523  & 0.578  & 0.529  & 0.547  & \textbf{0.508}  & 0.582  & 0.528 \\
\midrule

\multirow{5}{*}{\rotatebox{0}{$\text{ETTh2}$}}
& 96  & \cellcolor{red!10}\underline{0.230}  & \cellcolor{red!10}\underline{0.314}  & 0.239  & 0.325  & \cellcolor{red!10}\textbf{0.223}  & \cellcolor{red!10}\textbf{0.310}  & 0.231  & 0.318  & 0.235  & 0.320  & \underline{0.230}  & 0.316  & 0.260  & 0.332  & 0.244  & 0.324  & 0.252  & 0.330  & 0.245  & 0.325  & 0.298  & 0.369 \\
& 192 & \cellcolor{red!10}\underline{0.277}  & \cellcolor{red!10}0.350  & \cellcolor{red!10}0.280  & \cellcolor{red!10}0.352  & \cellcolor{red!10}\textbf{0.275}  & \cellcolor{red!10}\textbf{0.348}  & 0.287  & 0.358  & 0.293  & 0.359  & \underline{0.277}  & \underline{0.349}  & 0.287  & 0.359  & 0.301  & 0.364  & 0.308  & 0.368  & 0.299  & 0.364  & 0.350  & 0.402 \\
& 336 & \cellcolor{red!10}0.314  & \cellcolor{red!10}0.375  & \cellcolor{red!10}0.313  & \cellcolor{red!10}0.377  & \cellcolor{red!10}\textbf{0.309}  & \cellcolor{red!10}\textbf{0.371}  & 0.365  & 0.412  & 0.327  & 0.382  & \underline{0.311}  & \underline{0.373}  & 0.320  & 0.398  & 0.330  & 0.384  & 0.340  & 0.391  & 0.322  & 0.385  & 0.381  & 0.420 \\
& 720 & \cellcolor{red!10}0.389  & \cellcolor{red!10}0.421  & \cellcolor{red!10}0.388  & \cellcolor{red!10}0.421  & \cellcolor{red!10}\underline{0.381}  & \cellcolor{red!10}\underline{0.415}  & 0.410  & 0.435  & 0.412  & 0.432  & 0.392  & 0.422  & \textbf{0.333}  & \textbf{0.412}  & 0.392  & 0.421  & 0.395  & 0.424  & 0.399  & 0.447  & 0.418  & 0.436 \\
& Avg & \cellcolor{red!10}0.302  & \cellcolor{red!10}0.365  & \cellcolor{red!10}0.305  & \cellcolor{red!10}0.369  & \cellcolor{red!10}\textbf{0.297}  & \cellcolor{red!10}\textbf{0.361}  & 0.323  & 0.381  & 0.317  & 0.373  & 0.302  & \underline{0.365}  & \underline{0.300}  & 0.375  & 0.317  & 0.373  & 0.324  & 0.378  & 0.316  & 0.380  & 0.362  & 0.407 \\
\midrule

\multirow{5}{*}{\rotatebox{0}{$\text{ETTm1}$}}
& 96  & \cellcolor{red!10}0.370  & \cellcolor{red!10}0.380  & \cellcolor{red!10}0.367  & \cellcolor{red!10}0.383  & \cellcolor{red!10}0.380  & \cellcolor{red!10}0.378  & 0.384  & 0.385  & 0.417  & 0.401  & 0.381  & 0.380  & 0.457  & 0.444  & \underline{0.361}  & \underline{0.376}  & 0.369  & 0.377  & \textbf{0.356}  & \textbf{0.370}  & 0.412  & 0.424 \\
& 192 & \cellcolor{red!10}\textbf{0.431}  & \cellcolor{red!10}0.415  & \cellcolor{red!10}0.436  & \cellcolor{red!10}0.417  & \cellcolor{red!10}0.443  & \cellcolor{red!10}\underline{0.413}  & 0.445  & 0.423  & 0.457  & 0.429  & 0.446  & 0.417  & 0.473  & 0.454  & 0.435  & 0.413  & 0.441  & 0.415  & \underline{0.432}  & \textbf{0.408}  & 0.561  & 0.502 \\
& 336 & \cellcolor{red!10}\textbf{0.500}  & \cellcolor{red!10}0.456  & \cellcolor{red!10}0.523  & \cellcolor{red!10}0.470  & \cellcolor{red!10}0.515  & \cellcolor{red!10}\underline{0.452}  & 0.514  & 0.462  & 0.538  & 0.472  & 0.519  & 0.457  & 0.547  & 0.499  & 0.512  & 0.454  & 0.524  & 0.462  & \underline{0.502}  & \textbf{0.446}  & 0.598  & 0.513 \\
& 720 & \cellcolor{red!10}\underline{0.569}  & \cellcolor{red!10}0.506  & \cellcolor{red!10}0.586  & \cellcolor{red!10}0.509  & 0.585  & \cellcolor{red!10}\underline{0.501}  & 0.581  & 0.509  & 0.608  & 0.517  & 0.580  & 0.507  & 0.576  & 0.524  & 0.581  & 0.503  & 0.580  & 0.503  & \textbf{0.561}  & \textbf{0.490}  & 0.831  & 0.642 \\
& Avg & \cellcolor{red!10}\underline{0.467}  & \cellcolor{red!10}0.439  & \cellcolor{red!10}0.478  & \cellcolor{red!10}0.445  & 0.481  & \cellcolor{red!10}\underline{0.436}  & 0.481  & 0.445  & 0.505  & 0.454  & 0.481  & 0.440  & 0.513  & 0.480  & 0.472  & 0.437  & 0.478  & 0.439  & \textbf{0.463}  & \textbf{0.428}  & 0.600  & 0.520 \\
\midrule

\multirow{5}{*}{\rotatebox{0}{$\text{ETTm2}$}}
& 96  & \cellcolor{red!10}\textbf{0.155}  & \underline{0.253}  & \cellcolor{red!10}0.158  & \cellcolor{red!10}0.254  & \cellcolor{red!10}0.158  & \cellcolor{red!10}0.254  & 0.157  & \underline{0.253}  & 0.159  & 0.256  & 0.159  & 0.255  & 0.162  & 0.265  & 0.158  & 0.254  & \underline{0.156}  & \textbf{0.252}  & 0.160  & 0.256  & 0.162  & 0.265 \\
& 192 & \underline{0.191}  & \cellcolor{red!10}\textbf{0.284}  & 0.192  & \cellcolor{red!10}\underline{0.285}  & \cellcolor{red!10}0.191  & \cellcolor{red!10}\textbf{0.284}  & 0.191  & 0.287  & 0.192  & 0.286  & 0.192  & \underline{0.285}  & \textbf{0.188}  & 0.290  & 0.192  & \underline{0.285}  & 0.194  & 0.286  & 0.192  & \underline{0.285}  & 0.202  & 0.301 \\
& 336 & 0.234  & 0.316  & \cellcolor{red!10}0.233  & \cellcolor{red!10}0.319  & \underline{0.230}  & \underline{0.314}  & 0.232  & 0.316  & 0.237  & 0.322  & \textbf{0.229}  & \textbf{0.313}  & 0.234  & 0.324  & 0.231  & 0.315  & 0.234  & 0.316  & 0.232  & 0.318  & 0.251  & 0.340 \\
& 720 & \cellcolor{red!10}\textbf{0.299}  & \cellcolor{red!10}\textbf{0.360}  & 0.307  & 0.364  & \textbf{0.299}  & \cellcolor{red!10}\textbf{0.360}  & 0.304  & 0.363  & 0.306  & 0.364  & \textbf{0.299}  & \underline{0.361}  & 0.332  & 0.395  & \underline{0.301}  & 0.362  & 0.302  & \underline{0.361}  & 0.300  & 0.367  & 0.323  & 0.380 \\
& Avg & \cellcolor{red!10}\textbf{0.220}  & \cellcolor{red!10}\textbf{0.303}  & \cellcolor{red!10}0.222  & \cellcolor{red!10}0.306  & \textbf{0.220}  & \cellcolor{red!10}\textbf{0.303}  & \underline{0.221}  & 0.305  & 0.223  & 0.307  & \textbf{0.220}  & \underline{0.304}  & 0.229  & 0.318  & \underline{0.221}  & \underline{0.304}  & 0.222  & \underline{0.304}  & \underline{0.221}  & 0.307  & 0.235  & 0.322 \\
\midrule

\multirow{5}{*}{\rotatebox{0}{$\text{Weather}$}}
& 96  & \cellcolor{red!10}0.183  & \cellcolor{red!10}\underline{0.216}  & \cellcolor{red!10}0.186  & \cellcolor{red!10}0.225  & \cellcolor{red!10}\underline{0.181}  & \textbf{0.215}  & 0.184  & 0.218  & 0.199  & 0.230  & 0.183  & \textbf{0.215}  & \textbf{0.175}  & 0.236  & 0.188  & 0.225  & 0.194  & 0.232  & 0.208  & 0.239  & 0.192  & 0.242 \\
& 192 & \cellcolor{red!10}0.233  & \cellcolor{red!10}\underline{0.260}  & \cellcolor{red!10}0.245  & 0.275  & 0.232  & \textbf{0.258}  & 0.235  & 0.262  & 0.253  & 0.275  & \underline{0.231}  & \textbf{0.258}  & \textbf{0.217}  & 0.277  & 0.233  & 0.263  & 0.240  & 0.271  & 0.247  & 0.274  & 0.268  & 0.315 \\
& 336 & 0.297  & 0.305  & \cellcolor{red!10}0.300  & 0.312  & \cellcolor{red!10}0.291  & \cellcolor{red!10}\textbf{0.301}  & 0.294  & \underline{0.303}  & 0.305  & 0.311  & 0.293  & 0.303  & \textbf{0.272}  & 0.318  & 0.291  & 0.306  & 0.295  & 0.309  & \underline{0.289}  & 0.314  & 0.295  & 0.319 \\
& 720 & \cellcolor{red!10}0.377  & \underline{0.358}  & \cellcolor{red!10}0.379  & \cellcolor{red!10}0.360  & \cellcolor{red!10}0.378  & \cellcolor{red!10}\textbf{0.356}  & 0.381  & \underline{0.358}  & 0.388  & 0.363  & 0.380  & \underline{0.358}  & \textbf{0.347}  & 0.376  & 0.381  & 0.359  & 0.382  & 0.363  & \underline{0.358}  & 0.364  & 0.393  & 0.378 \\
& Avg & \cellcolor{red!10}0.272  & 0.285  & \cellcolor{red!10}0.278  & \cellcolor{red!10}0.293  & \cellcolor{red!10}\underline{0.271}  & \cellcolor{red!10}\textbf{0.283}  & 0.273  & 0.285  & 0.286  & 0.295  & 0.272  & \underline{0.284}  & \textbf{0.253}  & 0.302  & 0.273  & 0.288  & 0.278  & 0.294  & 0.275  & 0.298  & 0.287  & 0.314 \\
\midrule

\multirow{5}{*}{\rotatebox{0}{$\text{ECL}$}}
& 96 & \underline{0.175}  & \cellcolor{red!10}\textbf{0.252}  & 0.184  & 0.255  & \cellcolor{red!10}0.180  & \cellcolor{red!10}0.255  & \textbf{0.174}  & 0.257  & 0.179  & \underline{0.253}  & 0.188  & 0.260  & 0.178  & 0.262  & 0.203  & 0.289  & 0.187  & 0.267  & 0.199  & 0.285  & 0.261  & 0.348 \\
& 192 & \underline{0.185}  & \cellcolor{red!10}\textbf{0.260}  & 0.201  & 0.279  & \cellcolor{red!10}0.186  & \cellcolor{red!10}0.264  & \textbf{0.182}  & 0.266  & 0.186  & \underline{0.261}  & 0.194  & 0.267  & 0.186  & 0.272  & 0.197  & 0.282  & 0.192  & 0.272  & 0.201  & 0.291  & 0.288  & 0.373 \\
& 336 & 0.201  & \cellcolor{red!10}0.277  & 0.210  & 0.286  & 0.203  & 0.280  & \underline{0.198}  & 0.282  & \textbf{0.196}  & \textbf{0.272}  & 0.202  & \underline{0.276}  & 0.205  & 0.288  & 0.221  & 0.303  & 0.226  & 0.303  & 0.218  & 0.306  & 0.326  & 0.398 \\
& 720 & 0.245  & \cellcolor{red!10}0.313  & 0.254  & 0.319  & \cellcolor{red!10}0.244  & 0.311  & \textbf{0.236}  & 0.314  & \underline{0.238}  & \textbf{0.307}  & 0.245  & \underline{0.310}  & 0.246  & 0.323  & 0.271  & 0.338  & 0.265  & 0.327  & 0.262  & 0.339  & 0.357  & 0.415 \\
& Avg & 0.201  & \cellcolor{red!10}\underline{0.276}  & 0.212  & 0.285  & \cellcolor{red!10}0.203  & \cellcolor{red!10}0.277  & \textbf{0.198}  & 0.280  & \underline{0.200}  & \textbf{0.273}  & 0.207  & 0.278  & 0.204  & 0.286  & 0.223  & 0.303  & 0.218  & 0.292  & 0.220  & 0.305  & 0.308  & 0.384 \\
\midrule

\multirow{5}{*}{\rotatebox{0}{$\text{Traffic}$}}
& 96 & 0.459  & 0.274  & 0.474  & 0.277  & 0.511  & 0.314  & \textbf{0.447}  & \textbf{0.252}  & \underline{0.453}  & \underline{0.257}  & 0.466  & 0.273  & 0.548  & 0.282  & 0.610  & 0.348  & 0.538  & 0.309  & 0.695  & 0.379  & 0.837  & 0.435 \\
& 192 & 0.474  & 0.280  & 0.490  & 0.286  & 0.512  & 0.312  & \textbf{0.458}  & \textbf{0.262}  & \underline{0.467}  & \underline{0.263}  & 0.481  & 0.279  & 0.562  & 0.287  & 0.588  & 0.336  & 0.521  & 0.300  & 0.630  & 0.353  & 0.891  & 0.459 \\
& 336 & \underline{0.490}  & \underline{0.285}  & \cellcolor{red!10}0.501  & 0.293  & 0.526  & 0.316  & \textbf{0.473}  & \textbf{0.266}  & 0.520  & 0.290  & 0.497  & 0.286  & 0.582  & 0.294  & 0.607  & 0.349  & 0.537  & 0.309  & 0.632  & 0.355  & 0.950  & 0.495 \\
& 720 & \underline{0.524}  & 0.305  & \cellcolor{red!10}0.537  & 0.312  & 0.556  & 0.332  & \textbf{0.502}  & \textbf{0.287}  & 0.555  & 0.306  & 0.530  & \underline{0.304}  & 0.626  & 0.312  & 0.639  & 0.363  & 0.555  & 0.320  & 0.673  & 0.378  & 0.869  & 0.478 \\
& Avg & \underline{0.487}  & 0.286  & 0.500  & 0.292  & 0.526  & 0.319  & \textbf{0.470}  & \textbf{0.267}  & 0.499  & \underline{0.279}  & 0.493  & 0.285  & 0.579  & 0.294  & 0.611  & 0.349  & 0.538  & 0.310  & 0.657  & 0.366  & 0.887  & 0.467 \\
\midrule

\multirow{5}{*}{\rotatebox{0}{$\text{Solar}$}}
& 96 & \cellcolor{red!10}0.211  & \cellcolor{red!10}0.228  & 0.232  & 0.240  & \cellcolor{red!10}0.224  & \cellcolor{red!10}\underline{0.225}  & 0.231  & 0.260  & \textbf{0.206} & \textbf{0.211}  & 0.228  & 0.236  & \underline{0.210}  & 0.285  & 0.258  & 0.263  & 0.302  & 0.289  & 0.277  & 0.287  & 0.278  & 0.321 \\
& 192 & \cellcolor{red!10}\underline{0.240}  & \cellcolor{red!10}\underline{0.244}  & 0.264  & 0.252  & 0.265  & \cellcolor{red!10}\underline{0.244}  & 0.264  & 0.272  & 0.244  & \textbf{0.233}  & 0.259  & 0.248  & \textbf{0.227}  & 0.292  & 0.303  & 0.286  & 0.345  & 0.310  & 0.310  & 0.305  & 0.345  & 0.389 \\
& 336 & \cellcolor{red!10}\underline{0.260}  & \cellcolor{red!10}\underline{0.260}  & 0.342  & 0.298  & \cellcolor{red!10}0.280  & \cellcolor{red!10}\textbf{0.257}  & 0.279  & 0.283  & 0.315  & 0.283  & 0.290  & 0.268  & \textbf{0.240}  & 0.298  & 0.346  & 0.310  & 0.374  & 0.325  & 0.354  & 0.320  & 0.341  & 0.378 \\
& 720 & 0.269  & 0.265  & \cellcolor{red!10}0.315  & \cellcolor{red!10}0.283  & 0.300  & \cellcolor{red!10}\underline{0.262}  & \underline{0.251}  & \textbf{0.255}  & 0.321  & 0.284  & 0.298  & 0.270  & \textbf{0.240}  & 0.290  & 0.356  & 0.316  & 0.396  & 0.339  & 0.366  & 0.316  & 0.384  & 0.444 \\
& Avg & \cellcolor{red!10}\underline{0.245}  & \cellcolor{red!10}\underline{0.249}  & 0.289  & 0.268  & \cellcolor{red!10}0.267  & \cellcolor{red!10}\textbf{0.247}  & 0.256  & 0.267  & 0.271  & 0.253  & 0.269  & 0.255  & \textbf{0.229}  & 0.291  & 0.316  & 0.294  & 0.354  & 0.316  & 0.327  & 0.307  & 0.337  & 0.383 \\

\bottomrule
\end{tabular}
}
\end{center}
\end{table*}

\subsection{Full Results of In-domain Forecast}
\label{app:full_results_in_domain}
To evaluate the effectiveness of the deep decoupling-based tokenization method in \model{}, we conduct extensive experiments on the eight public datasets described in Appendix~\ref{app:datasets} (with full results deferred to Table~\ref{tab:appendix:in_domain}).

\subsection{Full Results of Cross-domain Forecast (One to One)}
\label{app:full_results_one_to_one}
 We followed the zero-shot experimental setup of TimeMixer++~\cite{wang2024timemixer++} and conducted six zero-shot experiments based on four subsets of the ETT~\cite{haoyietal-informer-2021} (ETTh1, ETTh2, ETTm1, ETTm2). Table~\ref{tab:appendix:cross_domain_one_to_one} shows the Mean Squared Error ($\text{MSE}$) and Mean Absolute Error ($\text{MAE}$) metrics, along with the average performance, for \model{} and baseline models, across prediction lengths $T \in \{96, 192, 336, 720\}$.

\begin{table*}[h]
\centering
\caption{Full results of cross-domain (one to one) forecasting performance. Best results are highlighted in \textbf{bold}, and second best results are \underline{underlined}. Light red shading indicates improvements over the corresponding backbone.}
\label{tab:appendix:cross_domain_one_to_one}
\begin{center}
\resizebox{1\linewidth}{!}{
\begin{tabular}{cc||cccccc|cccccc|cc|cccccccc}
\toprule
\multicolumn{2}{c}{\textbf{Method}} & \multicolumn{2}{c}{\textbf{DFTNet(T)}} & \multicolumn{2}{c}{\textbf{DFTNet(G)}} & \multicolumn{2}{c}{\textbf{DFTNet(M)}} & \multicolumn{2}{c}{\textbf{PatchTST}} & \multicolumn{2}{c}{\textbf{GPT4TS}} & \multicolumn{2}{c}{\textbf{PatchMLP}} & \multicolumn{2}{c}{\textbf{TOTEM}} &
\multicolumn{2}{c}{\textbf{TimeMixer}} & \multicolumn{2}{c}{\textbf{TimesNet}} &
\multicolumn{2}{c}{\textbf{DLinear}} & \multicolumn{2}{c}{\textbf{FEDformer}}\\

 \cmidrule(lr){1-2} \cmidrule(lr){3-8} \cmidrule(lr){9-14} \cmidrule(lr){15-16}  \cmidrule(lr){17-24}
\multicolumn{2}{c}{\textbf{Metric}} & \multicolumn{1}{c}{MSE} & \multicolumn{1}{c}{MAE} & \multicolumn{1}{c}{MSE} & \multicolumn{1}{c}{MAE} & \multicolumn{1}{c}{MSE} & \multicolumn{1}{c}{MAE} & \multicolumn{1}{c}{MSE} & \multicolumn{1}{c}{MAE} & \multicolumn{1}{c}{MSE} & \multicolumn{1}{c}{MAE} & \multicolumn{1}{c}{MSE} & \multicolumn{1}{c}{MAE} & \multicolumn{1}{c}{MSE} & \multicolumn{1}{c}{MAE} & \multicolumn{1}{c}{MSE} & \multicolumn{1}{c}{MAE} & \multicolumn{1}{c}{MSE} & \multicolumn{1}{c}{MAE} & \multicolumn{1}{c}{MSE} & \multicolumn{1}{c}{MAE} & \multicolumn{1}{c}{MSE} & \multicolumn{1}{c}{MAE} \\
\midrule

\multirow{5}{*}{\rotatebox{0}{$\text{ETTh1}$} $\rightarrow$ \rotatebox{0}{$\text{ETTh2}$}}
& 96  & \cellcolor{red!10}\textbf{0.234}  & \cellcolor{red!10}\underline{0.320}  & \cellcolor{red!10}0.236  & \cellcolor{red!10}0.321  & \cellcolor{red!10}\underline{0.235}  & \cellcolor{red!10}\textbf{0.319}  & 0.256  & 0.335  & 0.244  & 0.327  & 0.239  & 0.322  & 0.240  & 0.334  & 0.334  & 0.336  & 0.393  & 0.378  & 0.355  & 0.372  & 0.413  & 0.409 \\
& 192 & \cellcolor{red!10}\underline{0.295}  & \cellcolor{red!10}\underline{0.360}  & \cellcolor{red!10}\underline{0.295}  & \cellcolor{red!10}\underline{0.360}  & \textbf{0.292}  & \textbf{0.357}  & 0.309  & 0.370  & 0.308  & 0.366  & \textbf{0.292}  & \textbf{0.357}  & 0.299  & 0.378  & 0.417  & 0.378  & 0.460  & 0.412  & 0.426  & 0.410  & 0.497  & 0.456 \\
& 336 & \cellcolor{red!10}0.341  & \cellcolor{red!10}0.390  & \cellcolor{red!10}\underline{0.336}  & \cellcolor{red!10}0.386  & \textbf{0.333}  & \cellcolor{red!10}\textbf{0.383}  & 0.345  & 0.396  & 0.352  & 0.394  & \textbf{0.333}  & \underline{0.385}  & 0.377  & 0.432  & 0.514  & 0.427  & 0.551  & 0.458  & 0.499  & 0.452  & 0.587  & 0.504 \\
& 720 & \cellcolor{red!10}\textbf{0.409}  & \cellcolor{red!10}\textbf{0.429}  & \cellcolor{red!10}0.416  & \cellcolor{red!10}0.433  & \underline{0.413}  & 0.432  & 0.426  & 0.441  & 0.436  & 0.441  & \textbf{0.409}  & \underline{0.431}  & 0.520  & 0.517  & 0.665  & 0.506  & 0.715  & 0.536  & 0.650  & 0.537  & 0.744  & 0.572 \\
& Avg & \cellcolor{red!10}\underline{0.320}  & \cellcolor{red!10}0.375  & \cellcolor{red!10}0.321  & \cellcolor{red!10}0.375  & \textbf{0.318}  & \cellcolor{red!10}\textbf{0.373}  & 0.334  & 0.386  & 0.335  & 0.382  & \textbf{0.318}  & \underline{0.374}  & 0.359  & 0.415  & 0.483  & 0.412  & 0.530  & 0.446  & 0.483  & 0.443  & 0.560  & 0.485 \\
\midrule

\multirow{5}{*}{\rotatebox{0}{$\text{ETTh1}$} $\rightarrow$ \rotatebox{0}{$\text{ETTm2}$}}
& 96  & \cellcolor{red!10}\underline{0.198}  & \underline{0.307}  & \cellcolor{red!10}\textbf{0.195}  & \cellcolor{red!10}\textbf{0.305}  & \cellcolor{red!10}0.203  & \cellcolor{red!10}0.310  & 0.200  & \underline{0.307}  & 0.203  & \underline{0.307}  & 0.221  & 0.324  & 0.203  & 0.317  & 0.237  & 0.314  & 0.249  & 0.312  & 0.238  & 0.324  & 0.261  & 0.332 \\
& 192 & \cellcolor{red!10}\textbf{0.221}  & \cellcolor{red!10}\textbf{0.324}  & \cellcolor{red!10}\underline{0.224}  & \cellcolor{red!10}\underline{0.327}  & \cellcolor{red!10}0.230  & \cellcolor{red!10}0.331  & 0.246  & 0.344  & 0.240  & 0.335  & 0.242  & 0.339  & 0.253  & 0.354  & 0.296  & 0.341  & 0.304  & 0.338  & 0.299  & 0.355  & 0.329  & 0.368 \\
& 336 & \cellcolor{red!10}\underline{0.261}  & \cellcolor{red!10}\underline{0.351}  & \cellcolor{red!10}\textbf{0.258}  & \cellcolor{red!10}\textbf{0.348}  & \cellcolor{red!10}\textbf{0.258}  & \cellcolor{red!10}\textbf{0.348}  & 0.298  & 0.378  & 0.279  & 0.362  & 0.272  & 0.358  & 0.391  & 0.434  & 0.383  & 0.382  & 0.381  & 0.374  & 0.369  & 0.396  & 0.402  & 0.403 \\
& 720 & \cellcolor{red!10}0.343  & \cellcolor{red!10}0.397  & \cellcolor{red!10}\textbf{0.333}  & \cellcolor{red!10}\textbf{0.391}  & \cellcolor{red!10}\underline{0.337}  & \cellcolor{red!10}\underline{0.394}  & 0.359  & 0.408  & 0.351  & 0.401  & 0.350  & 0.401  & 0.589  & 0.540  & 0.474  & 0.419  & 0.486  & 0.424  & 0.494  & 0.467  & 0.494  & 0.440 \\
& Avg & \cellcolor{red!10}\underline{0.256}  & \cellcolor{red!10}\underline{0.345}  & \cellcolor{red!10}\textbf{0.253}  & \cellcolor{red!10}\textbf{0.343}  & \cellcolor{red!10}0.257  & \cellcolor{red!10}0.346  & 0.275  & 0.359  & 0.268  & 0.351  & 0.271  & 0.356  & 0.359  & 0.411  & 0.348  & 0.364  & 0.355  & 0.362  & 0.350  & 0.386  & 0.372  & 0.386 \\
\midrule

\multirow{5}{*}{\rotatebox{0}{$\text{ETTh2}$} $\rightarrow$ \rotatebox{0}{$\text{ETTh1}$}}
& 96  & \cellcolor{red!10}\underline{0.516}  & \cellcolor{red!10}\underline{0.480}  & 0.563  & 0.521  & 0.525  & \cellcolor{red!10}0.481  & 0.584  & 0.528  & 0.538  & 0.488  & 0.520  & 0.495  & 0.911  & 0.636  & \textbf{0.490}  & \textbf{0.473}  & 0.661  & 0.576  & 0.626  & 0.549  & 0.680  & 0.591 \\
& 192 & \cellcolor{red!10}\underline{0.569}  & \cellcolor{red!10}\underline{0.522}  & \cellcolor{red!10}0.594  & \cellcolor{red!10}0.541  & 0.582  & \underline{0.522}  & 0.667  & 0.585  & 0.626  & 0.542  & \textbf{0.557}  & \textbf{0.521}  & 0.914  & 0.656  & 0.626  & 0.535  & 0.738  & 0.594  & 0.703  & 0.582  & 0.871  & 0.664 \\
& 336 & \cellcolor{red!10}0.645  & \cellcolor{red!10}0.562  & \cellcolor{red!10}0.663  & 0.586  & \underline{0.624}  & \cellcolor{red!10}\textbf{0.553}  & 1.047  & 0.721  & 0.689  & 0.583  & \textbf{0.612}  & \underline{0.558}  & 0.941  & 0.699  & 0.744  & 0.594  & 0.863  & 0.646  & 0.770  & 0.611  & 0.890  & 0.672 \\
& 720 & \cellcolor{red!10}0.845  & \cellcolor{red!10}0.675  & \cellcolor{red!10}0.836  & \cellcolor{red!10}0.675  & \cellcolor{red!10}\textbf{0.786}  & \cellcolor{red!10}\textbf{0.645}  & 0.895  & 0.708  & 0.876  & 0.685  & \underline{0.822}  & 0.672  & 0.912  & 0.718  & 1.011  & 0.710  & 1.104  & 0.736  & 0.863  & \underline{0.656}  & 1.549  & 0.916 \\
& Avg & \cellcolor{red!10}0.644  & \cellcolor{red!10}\underline{0.560}  & \cellcolor{red!10}0.664  & 0.581  & \underline{0.629}  & \cellcolor{red!10}\textbf{0.550}  & 0.798  & 0.635  & 0.682  & 0.574  & \textbf{0.628}  & 0.562  & 0.920  & 0.677  & 0.718  & 0.578  & 0.842  & 0.638  & 0.741  & 0.600  & 0.998  & 0.711 \\
\midrule

\multirow{5}{*}{\rotatebox{0}{$\text{ETTm1}$} $\rightarrow$ \rotatebox{0}{$\text{ETTh2}$}}
& 96  & 0.295  & 0.367  & \cellcolor{red!10}0.294  & \cellcolor{red!10}0.371  & 0.297  & \cellcolor{red!10}0.368  & \underline{0.287}  & \underline{0.362}  & 0.318  & 0.382  & 0.297  & 0.369  & \textbf{0.271}  & \textbf{0.361}  & 0.413  & 0.386  & 0.453  & 0.423  & 0.401  & 0.413  & 0.452  & 0.445 \\
& 192 & 0.345  & \textbf{0.395}  & 0.360  & 0.410  & 0.357  & 0.400  & \underline{0.343}  & \textbf{0.395}  & 0.356  & 0.403  & 0.353  & \underline{0.398}  & \textbf{0.336}  & 0.401  & 0.489  & 0.420  & 0.613  & 0.505  & 0.484  & 0.453  & 0.935  & 0.708 \\
& 336 & 0.388  & 0.419  & \cellcolor{red!10}\underline{0.362}  & \cellcolor{red!10}\textbf{0.409}  & \cellcolor{red!10}0.385  & \cellcolor{red!10}0.415  & 0.375  & \underline{0.413}  & 0.390  & 0.422  & 0.386  & 0.418  & \textbf{0.356}  & 0.420  & 0.603  & 0.483  & 0.649  & 0.508  & 0.564  & 0.495  & 0.656  & 0.568 \\
& 720 & 0.481  & 0.471  & 0.492  & 0.482  & 0.465  & 0.458  & 0.462  & 0.463  & 0.473  & 0.466  & \underline{0.453}  & \underline{0.454}  & \textbf{0.388}  & \textbf{0.444}  & 0.756  & 0.558  & 0.823  & 0.588  & 0.696  & 0.558  & 0.838  & 0.652 \\
& Avg & 0.377  & 0.413  & \cellcolor{red!10}0.377  & 0.418  & 0.376  & 0.410  & \underline{0.367}  & \underline{0.408}  & 0.384  & 0.418  & 0.372  & 0.410  & \textbf{0.338}  & \textbf{0.406}  & 0.565  & 0.462  & 0.635  & 0.506  & 0.536  & 0.480  & 0.720  & 0.593 \\
\midrule

\multirow{5}{*}{\rotatebox{0}{$\text{ETTm1}$} $\rightarrow$ \rotatebox{0}{$\text{ETTm2}$}}
& 96 & \cellcolor{red!10}\textbf{0.166}  & \cellcolor{red!10}\textbf{0.263}  & \cellcolor{red!10}0.171  & \cellcolor{red!10}0.268  & \cellcolor{red!10}0.171  & \cellcolor{red!10}0.266  & \underline{0.168}  & \underline{0.265}  & 0.174  & 0.270  & 0.172  & 0.269  & 0.170  & 0.277  & 0.210  & \textbf{0.263}  & 0.242  & 0.301  & 0.256  & 0.341  & 0.296  & 0.376 \\
& 192 & \cellcolor{red!10}\textbf{0.201}  & \cellcolor{red!10}\textbf{0.292}  & \cellcolor{red!10}\underline{0.203}  & \cellcolor{red!10}\underline{0.294}  & \cellcolor{red!10}0.205  & \cellcolor{red!10}0.295  & \underline{0.203}  & 0.295  & 0.207  & 0.297  & 0.206  & 0.296  & 0.215  & 0.314  & 0.276  & 0.301  & 0.338  & 0.352  & 0.322  & 0.376  & 0.373  & 0.416 \\
& 336 & \textbf{0.243}  & \textbf{0.324}  & \cellcolor{red!10}\underline{0.244}  & \cellcolor{red!10}\textbf{0.324}  & 0.247  & \underline{0.326}  & \textbf{0.243}  & \textbf{0.324}  & 0.248  & \underline{0.326}  & 0.245  & \underline{0.326}  & 0.257  & 0.354  & 0.350  & 0.341  & 0.401  & 0.377  & 0.406  & 0.422  & 0.441  & 0.442 \\
& 720 & \cellcolor{red!10}0.321  & 0.373  & \cellcolor{red!10}\underline{0.318}  & \cellcolor{red!10}\textbf{0.369}  & 0.323  & \underline{0.372}  & 0.322  & \underline{0.372}  & 0.324  & 0.373  & \textbf{0.316}  & \textbf{0.369}  & 0.326  & 0.399  & 0.463  & 0.400  & 0.517  & 0.433  & 0.502  & 0.469  & 0.548  & 0.491 \\
& Avg & \cellcolor{red!10}\textbf{0.233}  & \cellcolor{red!10}\textbf{0.313}  & \cellcolor{red!10}\underline{0.234}  & \cellcolor{red!10}\underline{0.314}  & 0.236  & \cellcolor{red!10}\underline{0.314}  & \underline{0.234}  & \underline{0.314}  & 0.238  & 0.317  & 0.235  & 0.315  & 0.242  & 0.336  & 0.325  & 0.326  & 0.375  & 0.366  & 0.372  & 0.402  & 0.415  & 0.431 \\
\midrule

\multirow{5}{*}{\rotatebox{0}{$\text{ETTm2}$} $\rightarrow$ \rotatebox{0}{$\text{ETTm1}$}}
& 96 & \cellcolor{red!10}0.500  & 0.455  & \cellcolor{red!10}\underline{0.490}  & \cellcolor{red!10}0.450  & \cellcolor{red!10}0.495  & \cellcolor{red!10}\underline{0.447}  & 0.507  & 0.447  & 0.528  & 0.470  & 0.507  & 0.459  & 0.983  & 0.606  & \textbf{0.436}  & \textbf{0.428}  & 0.852  & 0.604  & 0.599  & 0.508  & 0.696  & 0.590 \\
& 192 & \cellcolor{red!10}0.554  & \cellcolor{red!10}0.484  & 0.562  & 0.486  & \cellcolor{red!10}0.547  & \cellcolor{red!10}\underline{0.475}  & 0.561  & 0.491  & \underline{0.540}  & 0.483  & 0.571  & 0.493  & 0.920  & 0.616  & \textbf{0.466}  & \textbf{0.451}  & 0.880  & 0.628  & 0.647  & 0.535  & 0.752  & 0.614 \\
& 336 & 0.669  & 0.539  & \cellcolor{red!10}\underline{0.582}  & \cellcolor{red!10}\underline{0.508}  & 0.609  & \cellcolor{red!10}0.509  & 0.595  & 0.511  & 0.645  & 0.535  & 0.597  & 0.510  & 1.124  & 0.663  & \textbf{0.548}  & \textbf{0.490}  & 0.751  & 0.596  & 0.722  & 0.573  & 0.811  & 0.639 \\
& 720 & 0.694  & 0.573  & \cellcolor{red!10}\textbf{0.655}  & \cellcolor{red!10}\textbf{0.553}  & \cellcolor{red!10}0.699  & \cellcolor{red!10}0.569  & \underline{0.678}  & \underline{0.567}  & 0.722  & 0.587  & 0.720  & 0.584  & 1.074  & 0.673  & 0.723  & 0.583  & 0.945  & 0.676  & 0.785  & 0.604  & 0.912  & 0.682 \\
& Avg & 0.604  & 0.513  & \cellcolor{red!10}\underline{0.572}  & \cellcolor{red!10}\underline{0.499}  & \cellcolor{red!10}0.588  & \cellcolor{red!10}0.500  & 0.585  & 0.504  & 0.609  & 0.519  & 0.599  & 0.512  & 1.026  & 0.640  & \textbf{0.543}  & \textbf{0.488}  & 0.857  & 0.626  & 0.688  & 0.555  & 0.793  & 0.631 \\
 
\bottomrule
\end{tabular}
}
\end{center}
\end{table*}

\subsection{Full Results of Cross-domain Forecast (Multi to One).}
\label{app:full_results_multi_to_one}
In this task setting, we selected the ETT~\cite{haoyietal-informer-2021}, ECL, Weather, Traffic~\cite{Wu2021AutoformerDT}, and Solar~\cite{lai2018modeling} as datasets. When one of these datasets was designated as the generalization target, the remaining datasets were used for model training (including both the first and second stages). We report in Table~\ref{tab:appendix:cross_domain_multi_to_one} the Mean Squared Error (MSE) and Mean Absolute Error (MAE) metrics, along with the average performance, for \model{} and baseline models, across prediction lengths $T \in \{96, 192, 336, 720\}$. 

It's worth noting that we trained all supervised models using the above data settings, while the pre-trained base models (including Moirai, Chronos, and Time-MOE) were not retrained. Therefore, \model{} actually achieved zero-shot prediction performance comparable to the base models by training on a small set of data.

\begin{table*}[h]
\centering
\caption{Full results of cross-domain (multi to one) forecasting performance. Best results are highlighted in \textbf{bold}, and second best results are \underline{underlined}. '-' denotes baselines pre-trained on the target domain and thus excluded from zero-shot comparison. Light red shading indicates improvements over the corresponding backbone.}
\label{tab:appendix:cross_domain_multi_to_one}
\begin{center}
\resizebox{1\linewidth}{!}{
\begin{tabular}{cc||cccccc|cccccc|cc|cccccc}
\toprule
\multicolumn{2}{c}{\textbf{Method}} & \multicolumn{2}{c}{\textbf{DFTNet(T)}} & \multicolumn{2}{c}{\textbf{DFTNet(G)}} & \multicolumn{2}{c}{\textbf{DFTNet(M)}} & \multicolumn{2}{c}{\textbf{PatchTST}} & \multicolumn{2}{c}{\textbf{GPT4TS}} & \multicolumn{2}{c}{\textbf{PatchMLP}} & \multicolumn{2}{c}{\textbf{TOTEM}} &
\multicolumn{2}{c}{\(\textbf{Moirai}_{base}\)} & \multicolumn{2}{c}{\(\textbf{Chronos}_{base}\)} &
\multicolumn{2}{c}{\(\textbf{Time-MOE}_{base}\)}\\

\cmidrule(lr){1-2} \cmidrule(lr){3-8} \cmidrule(lr){9-14} \cmidrule(lr){15-16}  \cmidrule(lr){17-22}

\multicolumn{2}{c}{\textbf{Metric}} & \multicolumn{1}{c}{MSE} & \multicolumn{1}{c}{MAE} & \multicolumn{1}{c}{MSE} & \multicolumn{1}{c}{MAE} & \multicolumn{1}{c}{MSE} & \multicolumn{1}{c}{MAE} & \multicolumn{1}{c}{MSE} & \multicolumn{1}{c}{MAE} & \multicolumn{1}{c}{MSE} & \multicolumn{1}{c}{MAE} & \multicolumn{1}{c}{MSE} & \multicolumn{1}{c}{MAE} & \multicolumn{1}{c}{MSE} & \multicolumn{1}{c}{MAE} & \multicolumn{1}{c}{MSE} & \multicolumn{1}{c}{MAE} & \multicolumn{1}{c}{MSE} & \multicolumn{1}{c}{MAE} & \multicolumn{1}{c}{MSE} & \multicolumn{1}{c}{MAE} \\
\midrule

\multirow{5}{*}{\rotatebox{0}{$\text{Multi}$} $\rightarrow$ \rotatebox{0}{$\text{Traffic}$}}
& 96  & \cellcolor{red!10}\textbf{0.739}  & \cellcolor{red!10}\underline{0.415}  & 0.897  & 0.520  & 0.785  & \cellcolor{red!10}0.430  & \underline{0.762}  & 0.435  & 0.850  & 0.484  & 0.779  & 0.431  & 0.866  & 0.500  &  -  &  -  & 0.843  & \textbf{0.335}  &  -  &  - \\
& 192 & \cellcolor{red!10}\textbf{0.676}  & \cellcolor{red!10}\underline{0.392}  & 0.835  & 0.504  & \cellcolor{red!10}\underline{0.708}  & \cellcolor{red!10}0.403  & 0.713  & 0.417  & 0.763  & 0.455  & 0.715  & 0.409  & 0.798  & 0.474  &  -  &  -  & 0.724  & \textbf{0.323}  &  -  &  - \\
& 336 & \cellcolor{red!10}\textbf{0.685}  & \cellcolor{red!10}\underline{0.395}  & 0.838  & 0.501  & \cellcolor{red!10}\underline{0.703}  & \cellcolor{red!10}0.403  & 0.725  & 0.408  & 0.732  & 0.427  & 0.713  & 0.411  & 0.804  & 0.473  &  -  &  -  & 0.712  & \textbf{0.329}  &  -  &  - \\
& 720 & \cellcolor{red!10}\textbf{0.728}  & \cellcolor{red!10}\underline{0.414}  & 0.915  & 0.527  & \cellcolor{red!10}\underline{0.749}  & \cellcolor{red!10}0.426  & 0.763  & 0.443  & 0.836  & 0.480  & 0.754  & 0.429  & 0.847  & 0.487  &  -  &  -  & 0.758  & \textbf{0.352}  &  -  &  - \\
& Avg & \cellcolor{red!10}\textbf{0.707}  & \cellcolor{red!10}\underline{0.404}  & 0.871  & 0.513  & \cellcolor{red!10}\underline{0.736}  & \cellcolor{red!10}0.415  & 0.741  & 0.426  & 0.795  & 0.461  & 0.740  & 0.420  & 0.829  & 0.484  &  -  &  -  & 0.759  & \textbf{0.335}  &  -  &  - \\
\midrule

\multirow{5}{*}{\rotatebox{0}{$\text{Multi}$} $\rightarrow$ \rotatebox{0}{$\text{Weather}$}}
& 96  & 0.234  & 0.259  & \cellcolor{red!10}0.222  & \cellcolor{red!10}0.257  & 0.246  & \cellcolor{red!10}\underline{0.253}  & \underline{0.220}  & 0.253  & 0.234  & 0.259  & 0.246  & 0.258  & 0.260  & 0.332  & \textbf{0.207}  & \textbf{0.231}  & 0.348  & 0.272  & 0.240  & 0.280 \\
& 192 & \cellcolor{red!10}\underline{0.267}  & \underline{0.290}  & \cellcolor{red!10}0.267  & \cellcolor{red!10}0.292  & 0.307  & \cellcolor{red!10}0.295  & 0.272  & 0.290  & 0.271  & 0.293  & 0.298  & 0.296  & 0.311  & 0.363  & \textbf{0.253}  & \textbf{0.274}  & 0.428  & 0.325  & 0.336  & 0.360 \\
& 336 & \cellcolor{red!10}0.323  & \cellcolor{red!10}\underline{0.327}  & \cellcolor{red!10}\underline{0.320}  & \cellcolor{red!10}0.329  & 0.359  & \cellcolor{red!10}0.331  & 0.329  & 0.341  & 0.330  & 0.331  & 0.356  & 0.333  & 0.347  & 0.388  & \textbf{0.302}  & \textbf{0.308}  & 0.427  & 0.346  & 0.435  & 0.432 \\
& 720 & \cellcolor{red!10}0.406  & \cellcolor{red!10}0.378  & \cellcolor{red!10}\underline{0.401}  & \cellcolor{red!10}\underline{0.376}  & 0.439  & 0.383  & 0.409  & 0.379  & 0.430  & 0.387  & 0.436  & 0.381  & 0.412  & 0.428  & \textbf{0.375}  & \textbf{0.355}  & 0.522  & 0.402  & 0.604  & 0.538 \\
& Avg & \cellcolor{red!10}0.307  & \cellcolor{red!10}\underline{0.313}  & \cellcolor{red!10}\underline{0.303}  & \cellcolor{red!10}0.314  & 0.338  & \cellcolor{red!10}0.316  & 0.308  & 0.316  & 0.316  & 0.317  & 0.334  & 0.317  & 0.333  & 0.378  & \textbf{0.284}  & \textbf{0.292}  & 0.431  & 0.336  & 0.404  & 0.402 \\
\midrule

\multirow{5}{*}{\rotatebox{0}{$\text{Multi}$} $\rightarrow$ \rotatebox{0}{$\text{ECL}$}}
& 96  & \cellcolor{red!10}0.255  & \cellcolor{red!10}0.329  & 0.297  & 0.381  & \cellcolor{red!10}0.285  & \cellcolor{red!10}0.342  & 0.296  & 0.360  & 0.291  & 0.368  & 0.300  & 0.367  & 0.283  & 0.370  & \underline{0.222}  & \underline{0.290}  &  -  &  -  & \textbf{0.191}  & \textbf{0.282} \\
& 192 & \cellcolor{red!10}0.243  & \cellcolor{red!10}0.319  & \cellcolor{red!10}0.271  & \cellcolor{red!10}0.354  & 0.268  & \cellcolor{red!10}0.331  & 0.268  & 0.340  & 0.277  & 0.359  & 0.264  & 0.340  & 0.268  & \textbf{0.268}  & \underline{0.227}  & \underline{0.296}  &  -  &  -  & \textbf{0.208}  & 0.301 \\
& 336 & \cellcolor{red!10}0.265  & \cellcolor{red!10}0.340  & \cellcolor{red!10}0.312  & \cellcolor{red!10}0.386  & \cellcolor{red!10}0.290  & \cellcolor{red!10}0.345  & 0.298  & 0.363  & 0.320  & 0.396  & 0.291  & 0.359  & 0.289  & 0.376  & \underline{0.248}  & \textbf{0.314}  &  -  &  -  & \textbf{0.241}  & \underline{0.330} \\
& 720 & \cellcolor{red!10}\underline{0.314}  & \cellcolor{red!10}\underline{0.371}  & \cellcolor{red!10}0.352  & \cellcolor{red!10}0.408  & 0.348  & \cellcolor{red!10}0.380  & 0.341  & 0.382  & 0.361  & 0.411  & 0.339  & 0.382  & 0.324  & 0.401  & \textbf{0.307}  & \textbf{0.355}  &  -  &  -  & 0.341  & 0.396 \\
& Avg & \cellcolor{red!10}0.269  & \cellcolor{red!10}0.340  & \cellcolor{red!10}0.308  & \cellcolor{red!10}0.382  & 0.298  & \cellcolor{red!10}0.349  & 0.301  & 0.361  & 0.312  & 0.383  & 0.298  & 0.362  & 0.291  & 0.354  & \underline{0.251}  & \textbf{0.314}  &  -  &  -  & \textbf{0.245}  & \underline{0.327} \\

\bottomrule
\end{tabular}
}
\end{center}
\end{table*}

\section{Visualization}
\label{app:visualization}
\begin{figure}
    \centering
    \includegraphics[width=0.8\linewidth]{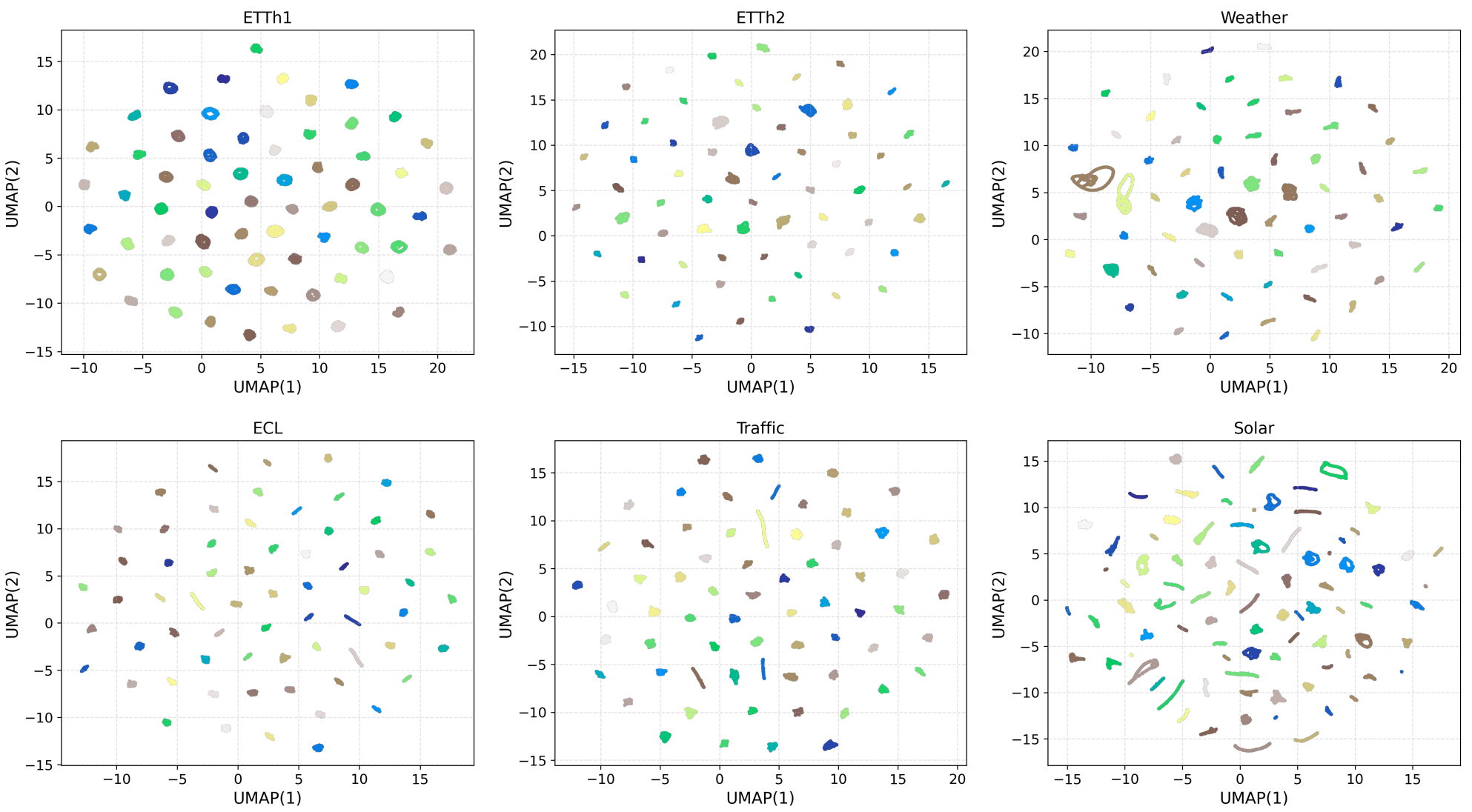}
    \caption{Dimensionality reduction visualization of the basis generated by DFT.}
    \label{fig:visualization_umap}
\end{figure}

We use UMAP to visualize the basis vectors output by the DFT module through dimensionality reduction. In our setup, the batch size is set to $B=512$ and the number of basis categories to $N=64$. Figure~\ref{fig:visualization_umap} shows the visualization results on the ETTh1, ETTh2, Weather, ECL, Traffic, and Solar datasets. Basis vectors of the same basis categories cluster into tight clusters due to consistency constraints, while different basis categories are subject to orthogonality constraints, resulting in clear boundary distances, thus achieving deep basis decouple of time series.

\section{Additional Ablation Studies}
\label{app:additional_ablation_study}

\subsection{Orthogonality Loss Weight \texorpdfstring{$\lambda$}{lambda} and Consistency Loss Weight \texorpdfstring{$\alpha$}{alpha}.}
\label{app:additional_ablation_study_loss_weight}
To control the contribution of orthogonality and consistency losses to the overall optimization objective, we used the hyperparameters \(\lambda\) and \(\alpha\) as weights for the two loss terms, respectively. Table~\ref{tab:hyper_study_loss_weight} reports the variations in reconstruction loss and codebook utilization of \model{} on ETTh1 with different \(\lambda\) and \(\alpha\) settings. The results show that reconstruction performance is optimal when both are set to 0.1, while codebook utilization reaches its highest level. When the weights are too large (e.g., 10), orthogonality and consistency losses often exceed the reconstruction loss in the later stages of training, thus interfering with the main optimization objective; conversely, when the weights are too small, the constraint effect is poor, and codebook utilization decreases accordingly.

\begin{table}[h]
\centering
\caption{Hyper-parameters study of orthogonality loss weight \(\lambda\) and consistency loss weight \(\alpha\).}
\small
\begin{tabular}{@{}ccccc@{}}
\toprule
Loss Weight \(\lambda / \alpha \) & MSE ($\downarrow$) & MAE ($\downarrow$) & Codebook Utilization(\%) ($\uparrow$) \\ 
\midrule
10 & 0.04386  & 0.14032  & 94.73 \\
1 & 0.00459  & 0.04682  & 98.06 \\
0.1 & \textbf{0.00414}  & \textbf{0.04353}  & \textbf{99.05} \\
0.01 & \underline{0.00422}  & \underline{0.04392}  & \underline{98.97} \\
0.001 & 0.00423  & 0.04398  & 98.96 \\
\bottomrule
\end{tabular}
\label{tab:hyper_study_loss_weight}
\end{table}

\subsection{The Order of Splitting and Encoding.} 
\label{app:additional_ablation_study_order_of_splitting_and_encoding}
This study employs a "Splitting after Encoding" strategy in obtaining individual tokens. The aim is to use the encoder to capture the complete information of a single sequence sample, and then decompose it into independent bases using the DFT module. Compared to the typical "Patch before encoding" strategy, this design ensures that the information contained in the obtained token is not limited by the length of a single patch, thus facilitating the capture of deeper information from the entire sequence sample. Table 4 shows a comparison of the performance of the two strategies on ETT1. The results show that the "Splitting after Encoding" design achieves superior performance across all prediction lengths on the ETT1 dataset, reducing the average MSE and average MAE by 6.81\% and 3.32\%, respectively.

\begin{table}[h]
\centering
\caption{The influence of the order of splitting and encoding.}
\small
\begin{tabular}{ccccc}
\toprule
\multicolumn{1}{c}{\textbf{Order}} & \multicolumn{2}{c}{\textbf{Splitting after Encoding}} & \multicolumn{2}{c}{\textbf{Splitting before Encoding}}\\
 \cmidrule(lr){1-1} \cmidrule(lr){2-3} \cmidrule(lr){4-5}
\multicolumn{1}{c}{\textbf{Metric}} & \multicolumn{1}{c}{MSE} & \multicolumn{1}{c}{MAE} & \multicolumn{1}{c}{MSE} & \multicolumn{1}{c}{MAE}\\
\midrule
96 & \textbf{0.433}  & \textbf{0.440}  & 0.468  & 0.452 \\
192 & \textbf{0.492}  & \textbf{0.479}  & 0.524  & 0.494 \\
336 & \textbf{0.542}  & \textbf{0.514}  & 0.615  & 0.554 \\
720 & \textbf{0.684}  & \textbf{0.603}  & 0.701  & 0.607 \\
Avg & \textbf{0.538}  & \textbf{0.509}  & 0.577  & 0.527 \\
\bottomrule
\end{tabular}
\label{tab:ablation_study_order_of_splitting_and_encoding}
\end{table}

\section{Efficiency Analysis}
\label{app:efficiency_analysis}
Table~\ref{tab:model_efficiency} shows the FLOPs estimation, runtime, and peak GPU memory usage statistics for \modelt{} and the baseline model (batch size = 512). All baseline models use the default recommended configuration. The results show that the computational cost of introducing \model{} only increases by 0.18\% compared to the original model (PatchTST) during the inference phase, with the main additional overhead concentrated in the first stage (tokenizer training).

\begin{table}[h]
\centering
\caption{Efficiency analysis.}
\small
\begin{tabular}{@{}lcccc@{}}
\toprule
Method & FLOPs (Millions) & Running Time (s/iter) & GPU Memory (MB) \\ 
\midrule
\modelt{} (stage 1) & 676188.914 & 0.051 & 21006.27 \\
\modelt{} (stage 2) & 39141.245 & 0.0524 & 22132.39 \\
PatchTST & 39069.942 & 0.0031 & 626.07 \\
FEDformer & 606395.744 & 0.0501 & 10300.32 \\
TimesNet & 290466.824 & 0.0862 & 2866.73 \\
TimeMixer & 786.727 & 0.0035 & 370.36 \\
Dlinear & 9.486 & 0.0001 & 191.28 \\
\bottomrule
\end{tabular}
\label{tab:model_efficiency}
\end{table}

\end{document}